\documentclass[10pt,twocolumn,letterpaper]{article}

\usepackage{cvpr}
\usepackage[accsupp]{axessibility}
\usepackage[dvipsnames]{xcolor}
\usepackage{graphicx}
\usepackage{booktabs}
\usepackage{colortbl}
\usepackage{array}
\usepackage[pagebackref,breaklinks,colorlinks,hidelinks,colorlinks=true,linkcolor=blue,citecolor=blue]{hyperref}
\usepackage[font=small,skip=5pt]{caption}
\usepackage{subcaption}
\usepackage{multirow}

\newcommand{\method}[1]{VisDiff}
\newcommand{\benchmark}[1]{VisDiffBench}
\newcommand{\ourbenchmark}[1]{PairedImageSets}
\newcommand{\Task}[1]{Set difference captioning}
\newcommand{\task}[1]{set difference captioning}

\newcommand{\seta}{$\mathcal{D}_A$}
\newcommand{\setb}{$\mathcal{D}_B$}
\newcommand{\subseta}{$\mathcal{S}_A$}
\newcommand{\subsetb}{$\mathcal{S}_B$}

\def\paperID{10346}
\def\confName{CVPR}
\def\confYear{2024}

\title{Describing Differences in Image Sets with Natural Language}

\author{Lisa Dunlap$^{*}$\\
UC Berkeley\\
{\tt\scriptsize \makebox[14em][c]{lisabdunlap@berkeley.edu}} \\
\and
Yuhui Zhang$^{*}$\\
Stanford\\
{\tt\scriptsize \makebox[14em][c]{yuhuiz@stanford.edu}} \\
\and
Xiaohan Wang\\
Stanford\\
{\tt\scriptsize \makebox[14em][c]{xhanwang@stanford.edu}} \\
\and
Ruiqi Zhong\\
UC Berkeley\\
{\tt\scriptsize \makebox[14em][c]{ruiqi-zhong@berkeley.edu}} \\
\and
Trevor Darrell$^{\dagger}$\\
UC Berkeley\\
{\tt\scriptsize \makebox[14em][c]{trevordarrell@berkeley.edu}} \\
\and
Jacob Steinhardt$^{\dagger}$\\
UC Berkeley\\
{\tt\scriptsize \makebox[14em][c]{jsteinhardt@berkeley.edu}} \\
\and
Joseph E. Gonzalez$^{\dagger}$\\
UC Berkeley\\
{\tt\scriptsize \makebox[14em][c]{jegonzal@berkeley.edu}} \\
\and
Serena Yeung-Levy$^{\dagger}$\\
Stanford\\
{\tt\scriptsize \makebox[14em][c]{syyeung@stanford.edu}} \\
}

\begin{document}

\twocolumn[{
\renewcommand\twocolumn[1][]{#1}
\maketitle
\begin{center}
    \vspace{-12mm}
    \centering
    \captionsetup{type=figure}
    \includegraphics[width=\textwidth,trim={0cm 3.5cm 0cm 0 cm},clip]{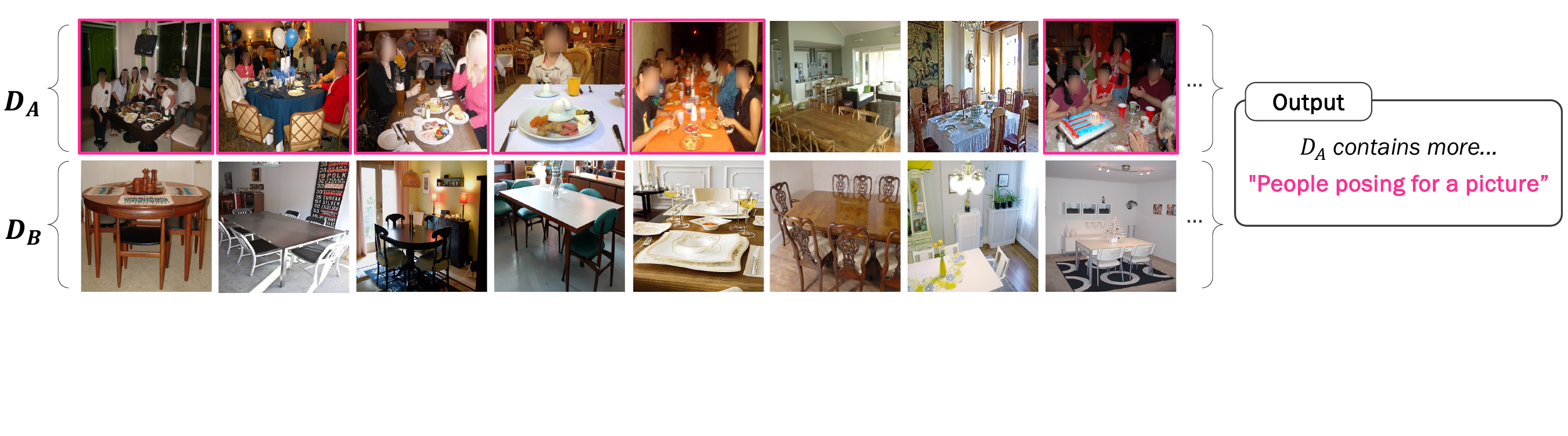}
    \captionof{figure}{\textbf{\Task{}.} Given two sets of images \seta{} and \setb{}, output natural language descriptions of concepts which are more true for \seta{}. In this example, \seta{} and \setb{} are images from the ``Dining Table'' class in ImageNetV2 and ImageNet, respectively. }
    \label{fig:teaser}
\end{center}
}]

{
  \renewcommand{\thefootnote}
    {\fnsymbol{footnote}}
  \footnotetext[1]{Equal contribution. $^\dagger$Equal advising. Both orders decided by coin flip.}
}

\begin{abstract}
\vspace{-3mm}
How do two sets of images differ? Discerning set-level differences is crucial for understanding model behaviors and analyzing datasets, yet manually sifting through thousands of images is impractical. To aid in this discovery process, we explore the task of automatically describing the differences between two \textbf{sets} of images, which we term Set Difference Captioning. This task takes in image sets $\mathcal{D}_A$ and $\mathcal{D}_B$, and outputs a description that is more often true on $\mathcal{D}_A$ than $\mathcal{D}_B$. We outline a two-stage approach that first proposes candidate difference descriptions from image sets and then re-ranks the candidates by checking how well they can differentiate the two sets. We introduce VisDiff, which first captions the images and prompts a language model to propose candidate descriptions, then re-ranks these descriptions using CLIP. To evaluate VisDiff, we collect VisDiffBench, a dataset with 187 paired image sets with ground truth difference descriptions. We apply VisDiff to various domains, such as comparing datasets (e.g., ImageNet vs. ImageNetV2), comparing classification models (e.g., zero-shot CLIP vs. supervised ResNet), characterizing differences between generative models (e.g., StableDiffusionV1 and V2), and discovering what makes images memorable. Using VisDiff, we are able to find interesting and previously unknown differences in datasets and models, demonstrating its utility in revealing nuanced insights.\footnote{Project page available at \url{https://understanding-visual-datasets.github.io/VisDiff-website/}.}
\vspace{-3mm}
\end{abstract}  
\begin{figure*}[t]
    \centering
    \includegraphics[width=\textwidth]{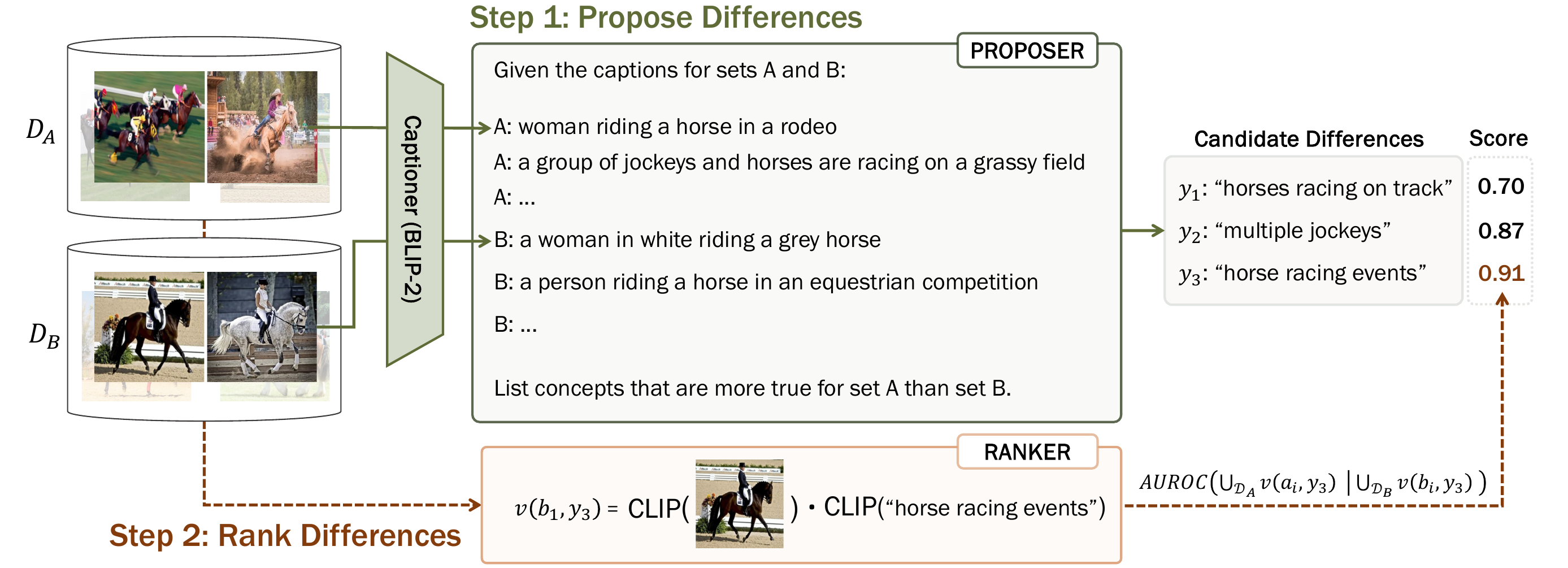}
    \caption{\textbf{\method{} algorithm.} VisDiff consists of a \emph{GPT-4 proposer} on \emph{BLIP-2} generated captions and a \emph{CLIP ranker}. The \emph{proposer} takes randomly sampled image captions from \seta{} and \setb{} and proposes candidate differences. The \emph{ranker} takes these proposed differences and evaluates them across all the images in \seta{} and \setb{} to assess which ones are most true.}
    \label{fig:main}
    \vspace{-3mm}
\end{figure*}

\section{Introduction}
\label{sec:intro}

What kinds of images are more likely to cause errors in one classifier versus another~\cite {eyuboglu2021domino, jain2023distilling}? 
How do visual concepts shift from a decade ago to now~\cite {quinonero2008dataset,zhu2022gsclip,koh2021wilds}?
What types of images are more or less memorable for humans~\citep{isola2011makes}?
Answering these questions can help us audit and improve machine learning systems, understand cultural changes, and gain insights into human cognition.

Although these questions have been independently studied in prior works, they all share a common desideratum: discovering differences between two sets of images. However, discovering differences in many, potentially very large, sets of images is a daunting task for humans. 
For example, one could gain insights into human memory by discovering systematic differences between memorable images and forgettable ones, but finding these differences may require scanning through thousands of images. 
An automated solution would be more scalable.

In this work, we explore the task of describing differences between image sets, which we term \emph{Set Difference Captioning}~(\autoref{fig:teaser}). 
Specifically, given two sets of images \seta{} and \setb{}, \task{} aims to find the most salient differences by generating natural language descriptions that are more often true in \seta{} than \setb{}.
We show in~\autoref{sec:applications} that many dataset and model analysis tasks 
can be formulated in terms of \task{}, and methods that address this problem can help humans discover new patterns in their data. 

\Task{} presents unique challenges to current machine learning systems, since it requires reasoning over all the given images.
However, no existing models in the vision and language space can effectively reason about thousands of images as input.
Furthermore, while there are usually many valid differences between \seta{} and \setb{}, end users are typically interested in what can most effectively differentiate between the two sets. For example, ``birthday party'' is a valid difference in~\autoref{fig:teaser}, but ``people posing for a picture'' better separates the sets. 

We introduce a two-stage proposer-ranker approach~\cite{zhongd3,zhongd5,zhu2022gsclip} for \task{} that addresses these challenges. 
As shown in~\autoref{fig:main},
the \emph{proposer} randomly samples subsets of images from $\mathcal{D}_A$ and $\mathcal{D}_B$ to generate a set of candidate differences in natural language. The \emph{ranker} then scores the salience and significance of each candidate by validating how often this difference is true for individual samples in the sets. 
Within the proposer-ranker framework, there are many plausible design choices for each component, and in this work we investigate three categories of proposers and rankers that utilize different combinations of models pre-trained with different objectives. 

To evaluate design choices, we construct \benchmark{} (\autoref{fig:visdiffbench_examples}), a dataset consisting of 187 paired image sets with ground-truth differences. We also propose a large language model-based evaluation to measure correctness. 
By benchmarking different designs on \benchmark{}, we identify our best algorithm, \method{}, which combines a proposer based on BLIP-2 captions and GPT-4 with a ranker based on CLIP features. This method accurately identifies 61\% and 80\% of differences using top-1 and top-5 evaluation even on the most challenging split of VisDiffBench.

Finally, we apply \method{} to a variety of applications, such as finding dataset differences, comparing model behaviors, and understanding questions in cognitive science. \method{} identifies both differences that can be validated by prior works, as well as new findings that may motivate future investigation. For example, \method{} uncovers ImageNetV2's temporal shift compared to ImageNet~\cite{recht2019imagenet,deng2009imagenet}, CLIP's strength in recognizing texts within images compared to ResNet~\cite{radford2021learning,he2016deep}, StableDiffusionV2 generated images' stylistic changes compared to StableDiffusionV1~\cite{rombach2022high}, and what images are more memorable by humans~\cite{IsolaParikhTorralbaOliva2011}.
These results indicate that the task of \task{} is automatic, versatile, and practically useful, opening up a wealth of new application opportunities for future work and potentially mass-producing insights unknown to even experts across a wide range of domains. 

\section{Related Works}

Many prior works explored difference captioning~\cite{li2023otter,li2023mimicit,Alayrac2022Flamingo,yao2022imagediff} and change captioning~\cite{robust_change_captioning,chg2cap,kim2021viewpoint}, which aim to describe differences between \emph{a single pair of images} with language. 
Recent large visual language models (VLMs) like GPT-4V~\cite{openai2023gpt4} have shown promise in describing differences in \emph{small groups of images}.
However, the question of how to scale this problem to sets containing \emph{thousands of images} remains unanswered. 
Meanwhile, some existing works in vision tackle understanding large amounts of visual data through finding concept-level prototypes~\cite{dataset_comparison,doersch2012what}, ``averaging'' large collections of images~\cite{zhu2014averageExplorer}, using simple methods like RGB value analysis~\cite{unbiased_look,Manovich2012}, or using a combination of detectors and classifiers to provide dataset level statistics~\cite{revisetool_eccv}. However, they do not describe the differences in natural language, which is flexible and easy to interpret.

Our work draws inspiration from D3~\cite{zhongd3} and D5~\cite{zhongd5} frameworks, which use large language models (LLMs) to describe differences between text datasets. A recent work GS-CLIP~\cite{zhu2022gsclip} applied a similar framework as D3 in the image domain, using CLIP features to retrieve differences from a pre-defined text bank. While this work targets the task of \task{}, it struggles at generating descriptive natural language and has a limited evaluation on the MetaShift~\cite{liang2021metashift} dataset that we found contains a significant amount of noise.
Inspired by D3~\cite{zhongd3}, our study advances a proposer-ranker framework tailored for the visual domain, leveraging large visual foundation models and a well-designed benchmark dataset. The versatility and effectiveness of our approach are further demonstrated through applications across a variety of real-world scenarios, underscoring its potential impact and utility in practical settings.

Lastly, the \task{} setting is closely related to the field of explainable computer vision.
Traditional explainable computer vision methodologies have predominantly concentrated on interpreting features or neurons within deep neural networks, as exemplified by approaches like LIME~\cite{ribeiro2016should}, CAM~\cite{zhou2016cvpr},  SHAP~\cite{lundberg2017unified}, and MILAN~\cite{hernandez2021natural}.
Recent shifts towards a data-centric AI paradigm have sparked a wave of research focusing on identifying influential data samples that impact predictions~\cite{park2023trak,shah2023modeldiff}, and on discerning interpretable data segments~\cite{chung2019automated,eyuboglu2021domino,deon2021spotlight}, thereby elucidating model behaviors. Our \task{} aligns with these objectives, offering a unique, purely data-driven approach to understanding and explaining differences in image sets with natural language. 

\section{Set Difference Captioning}
\label{sec:task}

In this section, we first describe the task of \task{}, then introduce VisDiffBench, which we use to benchmark performance on this task.

\begin{figure*}[!tb]
    \centering
    \includegraphics[width=\textwidth, ,trim={0cm 2.5cm 0cm 3cm},clip]{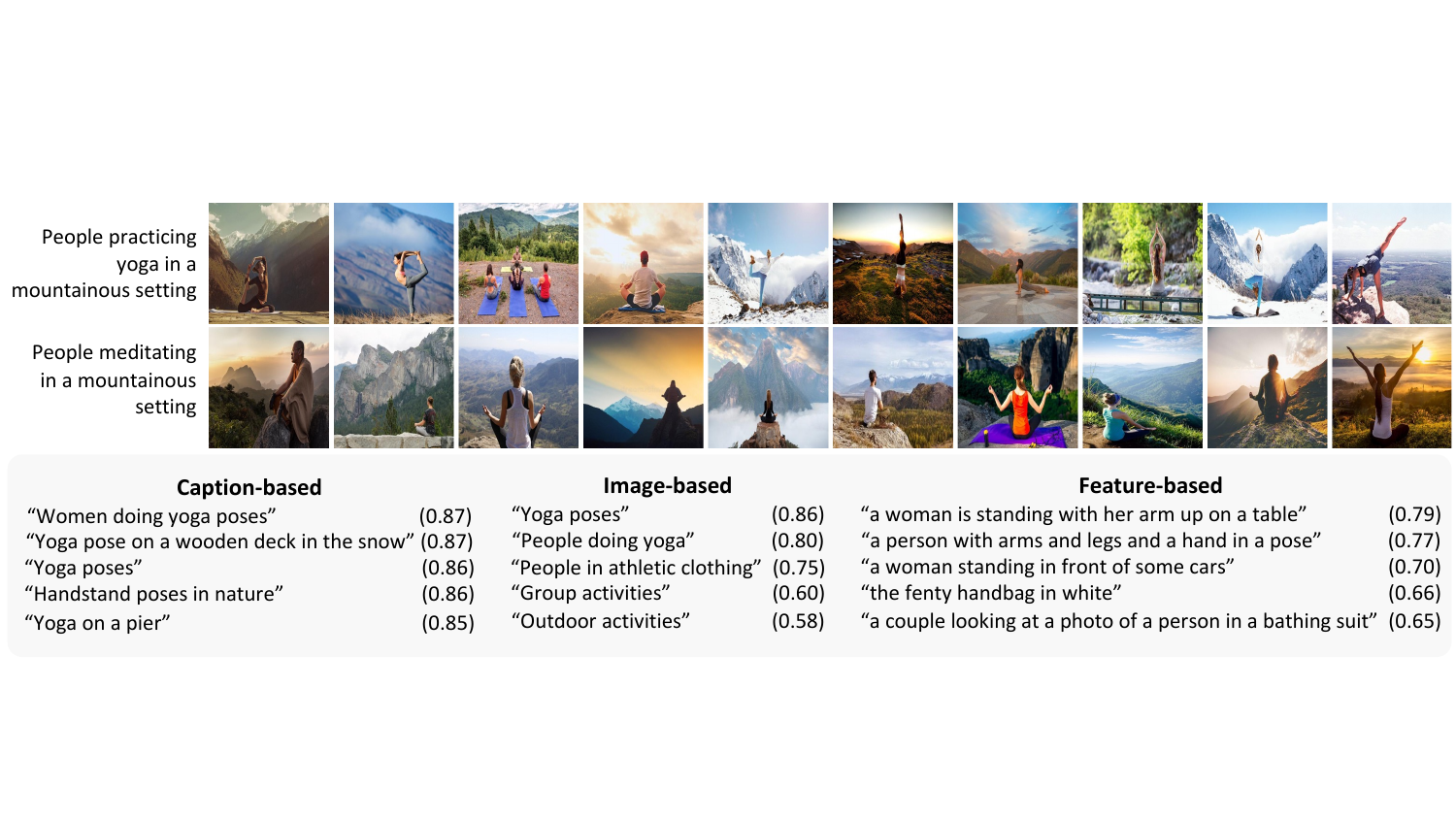}
    \vspace{-6mm}
    \caption{\textbf{Top 5 descriptions generated by the caption-based, image-based, and feature-based proposer.} All the top 5 descriptions from the caption-based proposer and the top 2 from the image-based proposer identify the ground-truth difference between ``practicing yoga'' and ``meditating'', while feature-based fails. We report AUROC scores from the same feature-based ranker described in~\autoref{sec:ranker-method}.}
    \label{fig:visdiffbench_examples}
    \vspace{-3mm}
\end{figure*}

\subsection{Task Definition}

Given two image datasets \seta{} and \setb{}, the goal of \emph{\task{} (SDC)} is to generate a natural language description $y$ that is more true in \seta{} compared to \setb{}. 
For example, in~\autoref{fig:visdiffbench_examples}, both \seta{} and \setb{} contain images of horses, but the images from \seta{} are all from racing events, so a valid choice of 
$y$ would be ``horse racing events''.

In our benchmarks below, we annotate (\seta{}, \setb{}) with a ground truth $y^*$ based on knowledge of the data-generating process. In these cases, we 
consider an output $y$ to be correct if it matches $y^*$ up to semantic equivalence (see~\autoref{sec:evaluation} for details). In our 
applications (\autoref{sec:applications}), we also consider cases where the ground truth $y^*$ is not known.

\subsection{Benchmark}
\label{sec:datasets}

To evaluate systems for \task{}, we construct \benchmark{}, a benchmark of 187 paired image sets each with a ground-truth difference description. 
To create \benchmark{}, we curated a dataset \ourbenchmark{} that covers 150 diverse real-world differences spanning three difficulty levels. We supplemented this with 37 differences obtained from two existing distribution shift benchmarks, ImageNet-R and ImageNet$^{*}$.  Aggregate statistics for \benchmark{} are given in~\autoref{tab:dataset_summary}.

\begin{table}[!tb]
\small
    \centering
    \begin{tabular}{lcc}
        \toprule
        \textbf{Dataset} & \textbf{\# Paired Sets} & \textbf{\# Images Per Set} \\ \midrule
        ImageNetR (sampled) & 14 & 500  \\
        ImageNet$^{*}$ (sampled) & 23 & 500 \\
        \ourbenchmark{} & \multirow{2}{*}{50/50/50} & \multirow{2}{*}{100/100/100} \\ 
        (Easy/Medium/Hard) &  &  \\ 
        \bottomrule
    \end{tabular}
    \caption{\textbf{Summary of \benchmark{}.} In experiments, we merge ImageNetR and ImageNet$^{*}$ because they have limited sets.}
    \label{tab:dataset_summary}
     \vspace{-3mm}
\end{table}

\textbf{ImageNet-R:} ImageNet-R~\cite{hendrycks2021many} contains renditions of 200 ImageNet classes across 14 categories (e.g., art, cartoon, painting, sculpture, sketch). For each category, we set $y^*$ to be the name of the category, \seta{} to be 500 images sampled from that category, and \setb{} to be 500 original ImageNet images sampled from the same 200 classes. 

\textbf{ImageNet$^{*}$:} ImageNet$^{*}$~\cite{vendrow2023dataset} contains 23 categories of synthetic images transformed from original ImageNet images using textual inversion. These categories include particular style, co-occurrence, weather, time of day, etc. For instance, one category, ``at dusk,'' converts ImageNet images with the prompt ``a photo of a [inverse image token] at dusk''. We generated differences analogously to ImageNet-R, taking \seta{} to be 500 image samples from the category and \setb{} to be 500 original ImageNet images. 

\textbf{\ourbenchmark{}:} 
ImageNetR and ImageNet$^{*}$ mainly capture stylistic differences, and only contain 37 differences in total. 
To address these shortcomings, we construct \textit{\ourbenchmark{}}, consisting of 150 paired image sets representing diverse differences. The dataset was built by first prompting GPT-4 to generate 150 paired sentences with three difficulty levels of differences (see~\autoref{sec:supp_sec3} for exact prompts). Easy level represents apparent difference (e.g., ``dogs playing in a park'' vs.~``cats playing in a park''), medium level represents fine-grained difference (e.g., ``SUVs on the road'' vs.~``sedans on the road''), and hard level represents subtle difference (e.g., ``people practicing yoga in a mountainous setting'' vs.~``people meditating in a mountainous setting''). 

Once GPT-4 generates the 150 paired sentences, we manually adjusted the annotated difficulty levels to match the criteria above. 
We then retrieved the top 100 images from Bing for each sentence. As a result, we collected 50 easy, 50 medium, and 50 hard paired image sets, with 100 images for each set.
One example pair from this dataset is shown in~\autoref{fig:visdiffbench_examples}, with further examples and a complete list of paired sentences provided in~\autoref{sec:supp_sec3}. 
We will release this dataset and the data collection pipeline.

\subsection{Evaluation}
\label{sec:evaluation}

To evaluate performance on \benchmark{}, we ask algorithms to output a description $y$ for each (\seta{}, \setb{}) pair and compare it to the ground truth $y^*$.
To automatically compute whether the proposed difference is semantically similar to the ground truth, we prompt GPT-4 to categorize similarity into three levels: 0 (no match), 0.5 (partially match), and 1 (perfect match); see~\autoref{sec:supp_sec3} for the exact prompt.

To validate this metric, we sampled 200 proposed differences on \ourbenchmark{} and computed the correlation of GPT-4's scores with the average score across four independent annotators.  
We observe a high Pearson correlation of 0.80, consistent with prior findings that large language models can align well with human evaluations~\citep{zheng2023judging,dubois2023alpacafarm}. 

We will evaluate systems that output ranked lists of proposals for each (\seta{}, \setb{}) pair. For these systems, we measure Acc@k, which is the highest score of any of the top-k proposals, 
averaged across all 187 paired image sets. 

\begin{table*}[t!]
    \small
    \centering
    \resizebox{\linewidth}{!}{
    \begin{tabular}{ll|cc|cc|cc|cc}
        \toprule
        \multirow{2}{*}{\textbf{Proposer}} & \multirow{2}{*}{\textbf{Ranker}} & \multicolumn{2}{c|}{\textbf{ImageNet-R/*}} & \multicolumn{2}{c|}{\textbf{PIS-Easy}} & \multicolumn{2}{c|}{\textbf{PIS-Medium}} & \multicolumn{2}{c}{\textbf{PIS-Hard}} \\ 
        & & Acc@1 & Acc@5 & Acc@1 & Acc@5 & Acc@1 & Acc@5 & Acc@1 & Acc@5 \\ \midrule
        Feature (BLIP-2) & Feature (CLIP) & 0.68 & 0.85 & 0.48 & 0.69 & 0.13 & 0.33 & 0.12 & 0.23 \\
        Image (LLaVA-1.5) & Feature (CLIP) & 0.27 & 0.39 & 0.71 & 0.81 & 0.39 & 0.49 & 0.28 & 0.43 \\
        Caption (BLIP-2 + GPT-4) & Caption (Vicuna-1.5) & 0.42 & 0.70 & 0.60 & 0.92 & 0.49 & 0.77 & 0.31 & 0.61 \\
        Caption (BLIP-2 + GPT-4) & Image (LLaVA-1.5) & 0.78 & 0.88 & 0.78 & \textbf{0.99} & 0.58 & 0.80 & 0.38 & 0.62 \\
        Image (GPT-4V) & Feature (CLIP) & \textbf{0.86} & 0.92 & \textbf{0.95} & \textbf{1.00} & \textbf{0.75} & \textbf{0.87} & 0.57 & 0.74 \\
        Caption (BLIP-2 + GPT-4) & Feature (CLIP) & 0.78 & \textbf{0.96} & 0.88 & \textbf{0.99} & \textbf{0.75} & \textbf{0.86} & \textbf{0.61} & \textbf{0.80} \\
        \bottomrule
    \end{tabular}
    }
    \caption{\textbf{Results on \benchmark{}.} GPT-4V image-based and BLIP-2 caption-based proposers with CLIP feature-based rankers consistently outperform other proposers and rankers by a large margin. We use the caption-based proposer with the CLIP ranker as the final \method{} algorithm because it obtains the highest accuracy on the \ourbenchmark{}-Hard and is cheaper than the GPT-4V image proposer. }
    \label{tab:results-modified}
    \vspace{-3mm}
\end{table*}

\label{sec:method}

\section{Our Method: \method{}}

It is challenging to train a neural network to directly predict $y$ based on \seta{} and \setb{}: \seta{} and \setb{} can be very large in practice, while currently no model can encode large sets of images and reliably reason over them. 
Therefore, we employ a two-stage framework for \task{}, using a {proposer} and a {ranker}~\cite{zhongd3,zhongd5}. The \emph{proposer} takes random subsets $\mathcal{S}_A \subseteq \mathcal{D}_A$ and $\mathcal{S}_B \subseteq \mathcal{D}_B$ and proposes differences. The \emph{ranker} takes these proposed differences and evaluates them across all of \seta{} and \setb{} to assess which ones are most true. We explore different choices of the proposer and ranker in the next two subsections. Full experiment details for this section, including the prompts for the models, can be found in~\autoref{sec:supp_sec4}.
 
\subsection{Proposer}

The proposer takes two subsets of images $\mathcal{S}_A$ and $\mathcal{S}_B$ as inputs and outputs a list $\mathcal{Y}$ of natural language descriptions that are (ideally) more true on $\mathcal{S}_A$ than $\mathcal{S}_B$. We leverage visual language models (VLM) as the proposer in three different ways: from the images directly, from the embeddings of the images, or by first captioning images and then using a language model. In all cases, we set $|\mathcal{S}_A|=|\mathcal{S}_B|=20$.

\textbf{Image-based Proposer:} We arrange the 20+20 input images into a single 4-row, 10-column grid and feed this as a single image into a VLM (in our case, LLaVA-1.5~\cite{liu2023visual} and GPT-4V~\cite{openai2023gpt4}). We then prompt the VLM to propose differences between the top and bottom half of images.

\textbf{Feature-based Proposer:} We embed images from \subseta{} and \subsetb{} into the VLM's visual representation space, then subtract the mean embeddings of \subseta{} and \subsetb{}. This subtracted embedding is fed into VLM's language model to generate a natural language description of the difference. We use BLIP-2~\cite{li2023blip} for this proposer.

\textbf{Caption-based Proposer:} We first use the VLM to generate captions of each image in \subseta{} and \subsetb{}. Then, we prompt a pure language model to generate proposed differences between the two sets of captions. We use BLIP-2 to generate the captions and GPT-4 to propose differences. 

Experiments in~\autoref{sec:proposer} show that the caption-based proposer works best, so we will take it as our main method and the other two as baselines. 
To further improve performance, we run the proposer multiple times over different sampled sets \subseta{} and \subsetb{}, then take the union of the proposed differences as inputs to the ranker.

\subsection{Ranker}
\label{sec:ranker-method}

Since the proposer operates on small  subsets \subseta{} and \subsetb{} and could generate invalid or noisy differences, we employ a \emph{ranker} to validate and rank the proposed differences $y \in \mathcal{Y}$. The ranker sorts hypotheses by computing a difference score $s_{y} = \mathbb{E}_{x \in \mathcal{D}_A}v(x,y) - \mathbb{E}_{x \in \mathcal{D}_B}v(x,y)$, where $v(x,y)$ is some measure of how well the image $x$ satisfies the hypothesis $y$. As before, we leverage VLMs to compute the ranking score $v(x,y)$ in three ways: from images directly, from image embeddings, and from image captions.

\textbf{Image-based Ranker:} We query the VQA model LLaVA-1.5~\cite{liu2023visual} to ask whether the image $x$ contains $y$, and set $v(x,y) = \text{VQA}(x,y)$ to be the resulting binary output.

\textbf{Caption-based Ranker:} We generate a caption $c$ from $x$ using BLIP-2~\cite{li2023blip}, then ask Vicuna-1.5~\cite{chiang2023vicuna} whether the caption $c$ contains $y$. We set $v(x,y) = \text{QA}(c,y)$ 
to be the resulting binary output.

\textbf{Feature-based Ranker:} We use CLIP ViT-G/14~\cite{radford2021learning} to compute the cosine similarity between the image embedding $e_x$ and text embedding $e_y$, so that $v(x,y) = \frac{e_x \cdot e_y}{\|e_x\|\|e_y\|}$. In contrast to the other two scores, since $v(x,y)$ is continuous rather than binary, we compute $s_y$ as the AUROC of using $v$ to classify between \seta{} and \setb{}.

Experiments in~\autoref{sec:ranker} show that the feature-based ranker achieves the best performance and efficiency, so we use it as our main method and the other two as baselines.
We also filter out proposed differences that are not statistically significant, by running a t-test on the two score distributions $v(x,y)$ with significance threshold $0.05$. 

\section{Results}
\label{sec:results}

In this section, we present experimental results to understand 1) which proposer / ranker works best, 2) can our algorithm consistently find the ground truth difference, and 3) can our algorithm work under noisy settings.

\subsection{Which Proposer is Best?}
\label{sec:proposer}

Our comparative results, presented in~\autoref{tab:results-modified}, demonstrate that \emph{the caption-based proposer consistently outperforms its image-based and feature-based counterparts by a large margin} across all subsets of the \benchmark{}. This difference is particularly pronounced in the most challenging subset, \ourbenchmark{}-Hard. While the captioning process may result in some loss of information from the original images, the strong reasoning capabilities of large language models effectively compensate for this by identifying \emph{diverse} and \emph{nuanced} differences between image sets. We provide a qualitative example in~\autoref{fig:visdiffbench_examples}.

The LLaVA image-based proposer shows commendable performance on \ourbenchmark{}-Easy but significantly lags behind the caption-based proposer on the \ourbenchmark{}-Medium/Hard subsets. Similarly, GPT-4V outperforms the caption-based proposer on the easy subset but underperforms on the hard subset. This discrepancy can be attributed to the loss of visual details when aggregating numerous images into a single gridded super-image. 

The feature-based proposer outperforms the LLaVA image-based proposer on ImageNetR and ImageNet$^{*}$ but is much less effective across all subsets of \ourbenchmark{}. We believe this is because the feature-based approach excels at distinguishing groups when one possesses attributes absent in the other (e.g., ``clipart of an image'' minus ``an image'' equates to ``clipart''). Most cases in ImageNetR/ImageNet$^{*}$ fit this scenario. However, this approach falls short in other situations where vector arithmetic does not yield meaningful semantic differences (e.g., ``cat'' minus ``dog'' is not semantically meaningful), which is a common scenario in \ourbenchmark{}.

\subsection{Which Ranker is Best?}
\label{sec:ranker}

In~\autoref{tab:results-modified}, our results demonstrate that \emph{the feature-based ranker consistently outperforms both the caption-based and image-based rankers}, particularly in the most challenging subset, \ourbenchmark{}-Hard. The feature-based approach's advantage is primarily due to its continuous scoring mechanism, which contrasts with the binary scores output by image-based and caption-based question answering. This continuous scoring allows for more fine-grained image annotation and improved calibration. It is also logical to observe the image-based ranker outperforms the caption-based one, as answering questions from original images tends to be more precise than from image captions.

Moreover, the efficiency of the feature-based ranker is remarkable. In scenarios where $M$ hypotheses are evaluated on $N$ images with $N \gg M$, the computation of image features is required only once. This results in a computational complexity of $O(M+N)\approx O(N)$, compared to $O(MN)$ for both image-based and caption-based rankers. Hence, the feature-based ranker requires significantly less computation, especially when ranking many hypotheses. This efficiency is crucial in practical applications, as we have found that a higher volume of proposed differences is essential for accurately identifying correct differences in the~\autoref{sec:supp_sec5}.

\subsection{Can Algorithm Find True Difference?}

In~\autoref{tab:results-modified}, the results demonstrate the effectiveness of our algorithm in discerning differences. The best algorithm, comprising a GPT-4~\cite{openai2023gpt4} caption-based proposer and a CLIP~\cite{radford2021learning} feature-based ranker, achieves accuracies of 88\%, 75\%, and 61\% for Acc@1, and 99\%, 86\%, and 80\% for Acc@5 on the PairedImageData-Easy/Medium/Hard subsets, respectively. The PairedImageData-Hard subset poses a significant challenge, requiring models to possess strong reasoning abilities to perceive extremely subtle variations, such as distinguishing between ``Fresh sushi with salmon topping'' and ``Fresh sushi with tuna topping'', or possess enough world knowledge to discern ``Men wearing Rolex watches'' from ``Men wearing Omega watches''. Despite these complexities, our model demonstrates impressive performance, accurately identifying specifics like ``Sushi with salmon'' and ``Men wearing Rolex watches''.

\subsection{Performance Under Noisy Data Splits}

In the \benchmark{} dataset, image sets are composed with perfect purity. For instance, \seta{} exclusively contains cat images (100\%), while \setb{} is entirely made up of dog images (100\%). However, this level of purity is rare in real-world scenarios. Typically, such sets include a mix of elements – for example, \seta{} might comprise 70\% cat images and 30\% dog images, and \setb{} vice versa. To evaluate the robustness of the \method{} algorithm against such noise, we introduced randomness in \benchmark{} by swapping a certain percentage of images between \seta{} and \setb{}. Here, 0\% purity signifies 50\% image swapping and an equal distribution of two sets, whereas 100\% purity indicates no image swapping.

\begin{figure}[!tb]
    \centering
    \includegraphics[width=\linewidth]{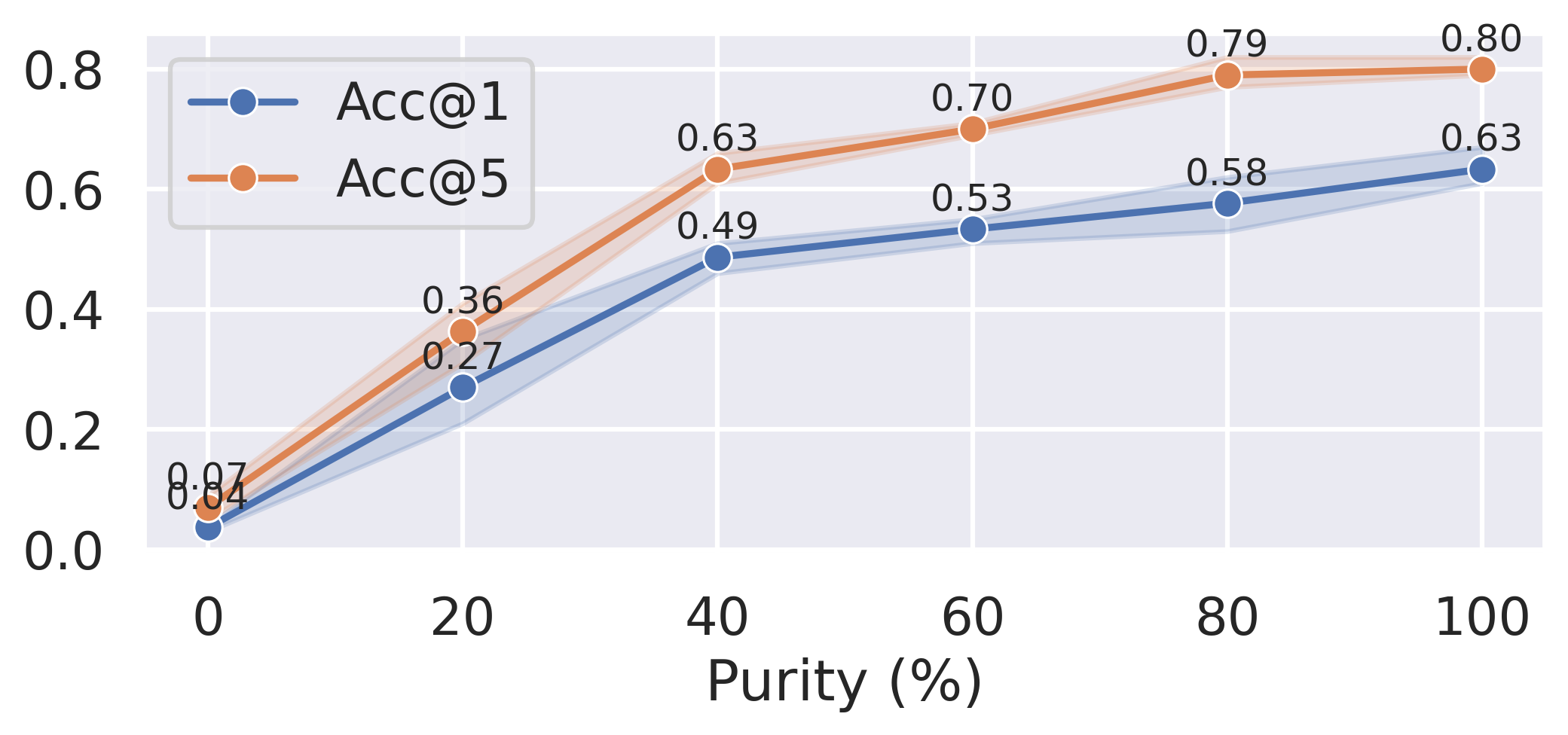}
    \caption{\textbf{\method{} performance under noise.} We randomly swap different percentages of images between \seta{} and \setb{} to inject noise. Results are computed on 50 paired sets in \ourbenchmark{}-Hard. 95\% confidence intervals are reported over three runs.}
    \vspace{-3mm}
    \label{fig:purity_analysis}
\end{figure}

\autoref{fig:purity_analysis} presents the Acc@1 and Acc@5 performance of \method{} across various purity levels, tested on 50 paired sets within \ourbenchmark{}-Hard. As anticipated, a decline in purity correlates with a drop in accuracy since identifying the difference becomes harder. However, even at 40\% purity, Acc@1 remains at 49\%, only modestly reduced from 63\% at 100\% purity. This result underscores the robustness of the \method{} algorithm to noisy data. It is also worth noting that \method{} reaches near 0\% accuracy at 0\% purity, which is expected since the two sets have exactly the same distribution and our method filters out invalid differences.

\paragraph{Other ablations of \method{} algorithm.} In~\autoref{sec:supp_sec5}, we further discuss how caption style, language model, sample size, and \# sampling rounds affect \method{} performance.

\section{Applications}
\label{sec:applications}

We apply the best configuration of our \method{} method to a set of five applications in computer vision: 1) comparing ImageNet and ImageNetV2 (\autoref{sec:imagenetv1v2}), 2) interpreting the differences between two classifiers at the datapoint level (\autoref{sec:clip_vs_resnet}), 3) analyzing model errors (\autoref{sec:auto-err-analysis}), 4) understanding the distributional output differences between StableDiffusionV1 and V2 (\autoref{sec:diffusion}), and 5) discovering what makes an image memorable (\autoref{sec:memorable}). 
Since \method{} is automatic, we used it to discover differences between (1) large sets of images or (2) many sets of images, thus mass-producing human-interpretable insights across these applications.
In this section, we report \method{}-generated insights including some that can be confirmed with existing work and others that may motivate future investigation in the community.
Additional details for each application can be found in~\autoref{sec:supp_sec6}.

\subsection{Comparing ImageNetV2 with ImageNet}
\label{sec:imagenetv1v2}

In 2019, a decade after ImageNet~\cite{deng2009imagenet} was collected, Recht et al. introduced ImageNetV2~\cite{recht2019imagenet}, which attempted to mirror the original ImageNet collection process, including restricting data to images uploaded in a similar timeframe. 
However, models trained on ImageNet showed a consistent 11-14\% accuracy drop on ImageNetV2, and the reasons for this have remained unclear. 
While some studies have employed statistical tools to reveal a distributional difference between ImageNet and ImageNetV2~\cite{engstrom2020identifying}, we aim to discover more interpretable differences between these two datasets. 

To uncover their differences, we first ran \method{} with \seta{} as all of ImageNetV2 images and \setb{} as all of ImageNet images.
Interestingly, the highest scoring description generated by our system is ``photos taken from Instagram''.  We conjecture that this highlights temporal distribution shift as a potential reason behind model performance drops on ImageNetV2 vs V1. Indeed, while ImageNetV2 aimed to curate images uploaded in a similar timeframe to ImageNet, all images in ImageNet were collected prior to 2012 whereas a portion of ImageNetV2 was collected between 2012 and 2014~\citep{recht2019imagenet}. This shift in years happens to coincide with the explosion of social media platforms such as Instagram, which grew from 50M users in 2012 to 300M users in 2014~\citep{Digital-2018}. In this case, we hypothesize that a small difference in the time range had a potentially outsized impact on the prevalence of Instagram-style photos in ImageNetV2 and the performance of models on this dataset. 

Beyond dataset-level analysis, we applied \method{} to each of the 1,000 ImageNet classes, comparing ImageNetV2 images (\seta{}) against ImageNet images (\setb{}). Notable class-specific differences are listed in~\autoref{tab:imagenetv2_example_classes}, ranked by difference score, with visualizations in~\autoref{supp_fig:imagenet_v2_examples}. Several of these differences suggest more specific examples of Instagram-style photos, consistent with our dataset-level finding. For example, for the class ``Dining Table", ImageNetV2 contains substantially more images showing ``people posing for a picture'', visualized in~\autoref{fig:teaser}. For the class ``Horizontal Bar", ImageNetV2 is also identified to have more images of ``men's gymnastics events." Upon manual inspection, we find that this highlights the difference that ImageNetV2 happens to contain photographs of the Men's High Bar gymnastics event in the 2012 Olympics, which occurred after the ImageNet collection date. These examples illustrate how VisDiff can be used as a tool for surfacing salient differences between datasets.

\begin{table}[!tb]
\scriptsize
\centering
\begin{tabular}{p{0.4\linewidth} p{0.5\linewidth}}
\toprule
\textbf{Class} & \textbf{More True for ImageNetV2}  \\
\midrule
Dining Table & People posing for a picture \\
Wig & Close up views of dolls \\
Hand-held Computer & Apps like Twitter and Whatsapp \\
Palace & East Asian architecture \\
Pier & Body of water at night \\
\bottomrule
\end{tabular}
\caption{\textbf{Top per-class differences between ImageNet and V2.}}
\label{tab:imagenetv2_example_classes}
\vspace{-3mm}
\end{table}

\subsection{Comparing Behaviors of CLIP and ResNet}
\label{sec:clip_vs_resnet}

In 2021, OpenAI's CLIP \citep{radford2021learning} showcased impressive zero-shot object recognition, matching the fully supervised ResNet~\cite{he2016deep} in ImageNet accuracy while showing a smaller performance drop on ImageNetV2. Despite similar in-distribution performance on ImageNet, CLIP and ResNet differ in robustness~\cite{miller2021accuracy}. This naturally leads to two questions: 1) do these models make similar predictions on individual datapoints in ImageNet? 2) on what datapoints does CLIP perform better than ResNet in ImageNetV2? 

To investigate these questions, we analyzed ResNet-50 and zero-shot CLIP ViT-H, which achieve similar accuracies of 75\% and 72\% on ImageNet, respectively.
To study the first question, \method{} was applied to the top 100 classes where CLIP surpasses ResNet. \seta{} comprised images correctly identified by CLIP but not by ResNet, and \setb{} included all other images.
The top discoveries included ``close-ups of everyday objects'', ``brands and specific product labeling'', and ``people interacting with objects''. 
The first two align well with existing works that show CLIP is robust to object angles and sensitive to textual elements (e.g., a fruit apple with text ``iPod'' on it will be misclassified as ``iPod'')~\cite{radford2021learning, goh2021multimodal}. 
In addition, we ran \method{} at finer granularity on each of the top 5 classes where CLIP outperforms ResNet. The discovered class-level differences are shown in~\autoref{tab:modeldiff_example_classes}, demonstrating CLIP's proficiency in identifying ``tobacco shops with signs'', ``group displays of digital watches'', and ``scenes involving missiles and toyshops with human interactions'', which echos the dataset-level findings about label, object angle, and presence of people. 

\begin{table}[!tb]
\scriptsize
\centering
\begin{tabular}{llll}
\toprule
\textbf{Class} & \textbf{Acc$_C$} & \textbf{Acc$_R$} & \textbf{More Correct for CLIP}  \\
\midrule
Tobacco Shop & 0.96 & 0.50 & Sign hanging from the side of a building \\
Digital Watch & 0.88 & 0.52 & Watches displayed in a group \\
Missile & 0.78 & 0.42 & People posing with large missiles \\
Pot Pie & 0.98 & 0.66 & Comparison of food size to coins \\
Toyshop & 0.92 & 0.60 & People shopping in store \\
\bottomrule
\end{tabular}
\caption{\textbf{Top per-class differences between CLIP and ResNet.} Acc$_C$ and $Acc_R$ are accuracy of CLIP and ResNet, respectively.}
\label{tab:modeldiff_example_classes}
\vspace{-5.5mm}
\end{table}

To study the second question, we applied \method{} to ImageNetV2's top 100 classes where CLIP outperforms ResNet. We set \seta{} as images where CLIP is correct and ResNet is wrong, and \setb{} as the rest.
The top three differences are: 1) ``Interaction between humans and objects'', suggesting CLIP's robustness in classifying images with human presence; 2) ``Daytime outdoor environments'', indicating CLIP's temporal robustness; and 3) ``Group gatherings or social interactions'', which is similar to the first difference. 
These findings provide insight into CLIP's strengths versus ResNet on ImageNetV2, and are also consistent with the findings in \autoref{sec:imagenetv1v2} that ImageNetV2 contains more social media images with more presence of people.

\begin{figure*}
    \centering
    \includegraphics[width=0.95\textwidth,trim={0cm 0cm 0cm 0cm},clip]{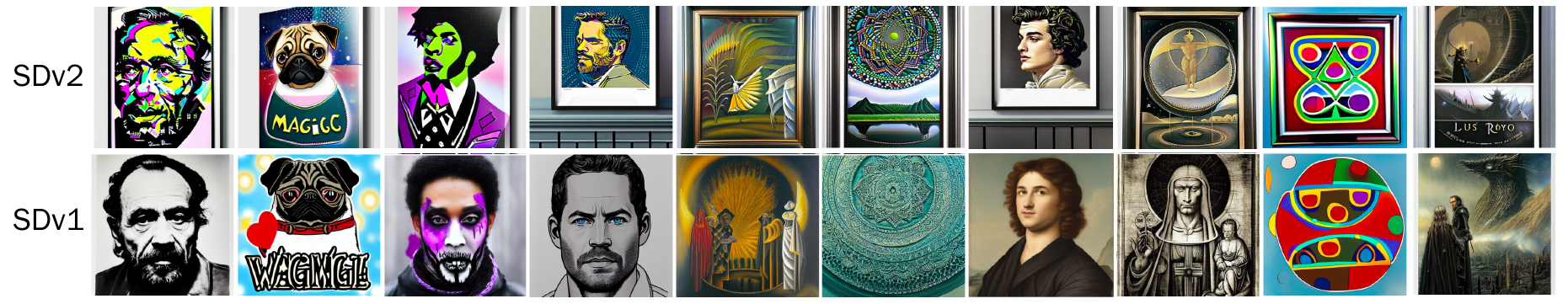}
    \caption{\textbf{StableDiffusionV2 vs. V1 generated images.} For the same prompt, StableDiffusionV2 images often contain more ``vibrant contrasting colors'' and ``artworks placed on stands or in frames''. Randomly sampled images can be found in~\autoref{supp_fig:diffusion_random_samples}.}
    \label{fig:diffusion_examples}
    \vspace{-5mm}
\end{figure*}

\subsection{Finding Failure Modes of ResNet} \label{sec:auto-err-analysis}

We utilize \method{} to identify failure modes of a model by contrasting images that are correctly predicted against those that are erroneously classified. Using a ResNet-50 and ResNet-101~\cite{he2016deep} trained on ImageNet, we set \seta{} as ImageNet images misclassified by both ResNet-50 and ResNet-101 and \setb{} as correctly classified images. 
The two highest scoring descriptions were ``humanized object items'' and ``people interacting with objects'', suggesting that ResNet models perform worse when the images include human subjects, echoing the finding in~\autoref{sec:clip_vs_resnet}. 

To validate this hypothesis, we applied a DETR~\cite{redmon2016you} object detector to find a subset of ImageNet images with human presence. Using the classes which have a roughly equal number of human/no-human images, we evaluated ResNets on this subset and their accuracy indeed declined 3-4\%, as shown in~\autoref{tab:resnet_failure}.

\begin{table}[!tb]
\scriptsize
\centering
\begin{tabular}{lcc}
\toprule
Model     & Images w/ Person & Images w/o Person \\ 
\midrule
ResNet50  & 67.24\%          & 69.96\%           \\
ResNet101 & 68.75\%          & 72.30\%           \\
Ensemble  & 74.86\%          & 77.32\%           \\ 
\bottomrule
\end{tabular}
\caption{\textbf{Accuracy on images with / without people.}}
\label{tab:resnet_failure}
\vspace{-3mm}
\end{table}


\subsection{Comparing Versions of Stable Diffusion} \label{sec:diffusion}

In 2022, Stability AI released StableDiffusionV1 (SDv1), followed by StableDiffusionV2 (SDv2)~\cite{rombach2022high}. While SDv2 can be seen as an update to SDv1, it raises the question: What are the differences in the images produced by these two models?

Using the prompts from PartiPrompts~\cite{parti} and  DiffusionDB~\cite{wangDiffusionDBLargescalePrompt2022}, we generated 1634 and 10000 images with SDv2 and SDv1, respectively. The Parti images are used to propose differences and the DiffusionDB images are used to validate these differences transfer to unseen prompts. 

The top differences show that SDv2 produces more ``vibrant and contrasting colors'' and interestingly ``images with frames or borders'' (see~\autoref{supp_tab:diffusion_hypotheses}). We confirmed the color difference quantitatively by computing the average saturation: 112.61 for SDv2 versus 110.45 for SDv1 from PartiPrompts, and 97.96 versus 93.49 on unseen DiffusionDB images. Qualitatively, as shown in Section~\autoref{fig:diffusion_examples}, SDv2 frequently produces images with white borders or frames, a previously unknown characteristic. This is further substantiated in Section~\autoref{sec:supp_sec6}, where we employ edge detection to quantify white borders, providing 50 random image samples from both SDv1 and SDv2.

\subsection{Describing Memorability in Images} 
\label{sec:memorable}

Finally, we demonstrate the applicability of \method{} in addressing diverse real-world questions beyond machine learning, such as computational cognitive science. A key area of interest, especially for photographers and advertisers, is enhancing image memorability. \citet{IsolaParikhTorralbaOliva2011} explored this question and created the LaMem dataset, where each image is assigned a memorability score by humans in the task of identifying repeated images in a sequence.

\begin{figure}
    \centering
    \includegraphics[width=0.48\textwidth]{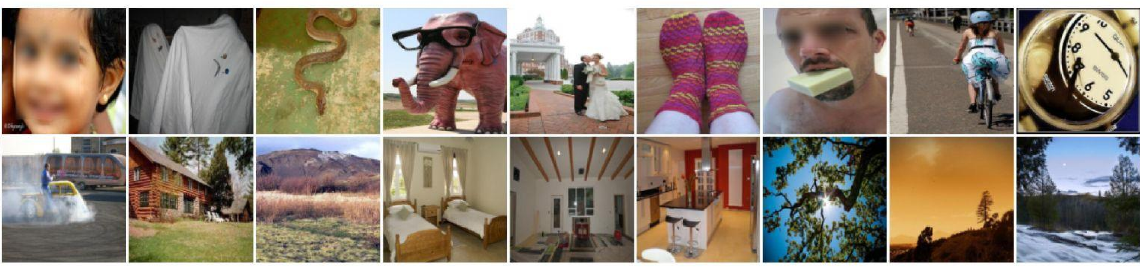}
    \caption{\textbf{Memorable(top) vs. forgettable(bottom) images.} Memorable images contain more ``humans'', ``close-up views of body part or objects'', and ``humorous settings'', while forgettable images contain more ``landscapes'' and ``urban environments''}
    \label{fig:lamem}
    \vspace{-3mm}
\end{figure}

Applying \method{} to the LaMem dataset, we divided images into two groups: \seta{} (the most memorable 25th percentile) and \setb{} (the least memorable 25th percentile). Our analysis found that memorable images often include ``presence of humans'', ``close-up views'', and ``humorous settings'', while forgettable ones feature ``landscapes'' and ``urban environments''. These findings are consistent with those of \citet{IsolaParikhTorralbaOliva2011}, as further detailed qualitatively in~\autoref{fig:lamem} and quantitatively in~\autoref{sec:supp_sec6}.

\section{Conclusion}
\label{sec:conclusion}

In this work, we introduce the task of \task{} and develop VisDiff, an algorithm designed to identify and describe differences in image sets in natural language. VisDiff first uses captioning and large language models to propose differences based on image captions and then employs CLIP to effectively rank these differences. We evaluate VisDiff's various design choices on our curated VisDiffBench, and show \method{}'s utility in finding interesting insights across a variety of real-world applications. 

\paragraph{Limitations.} 
As we see in Section \ref{sec:results}, \method{} still has a large room for improvement and hence far from guaranteed to uncover all meaningful differences.
Furthermore, \method{} is meant to be an assistive tool for humans to better understand their data and should not be applied without a human in the loop: the users hold the ultimate responsibility to interpret the descriptions by \method{} properly.
As \method{} relies heavily on CLIP, GPT, and BLIP, any biases or errors these models may extend to \method{}.
Further investigation of \method{}'s failure cases can be found in~\autoref{sec:supp_limitations}.

{
    \small
    \bibliographystyle{ieeenat_fullname}
    \bibliography{main}

\begin{thebibliography}{53}
\providecommand{\natexlab}[1]{#1}
\providecommand{\url}[1]{\texttt{#1}}
\expandafter\ifx\csname urlstyle\endcsname\relax
  \providecommand{\doi}[1]{doi: #1}\else
  \providecommand{\doi}{doi: \begingroup \urlstyle{rm}\Url}\fi

\bibitem[Alayrac et~al.(2022)Alayrac, Donahue, Luc, Miech, Barr, Hasson, Lenc, Mensch, Millican, Reynolds, et~al.]{Alayrac2022Flamingo}
Jean-Baptiste Alayrac, Jeff Donahue, Pauline Luc, Antoine Miech, Iain Barr, Yana Hasson, Karel Lenc, Arthur Mensch, Katherine Millican, Malcolm Reynolds, et~al.
\newblock Flamingo: a visual language model for few-shot learning.
\newblock In \emph{NeurIPS}, 2022.

\bibitem[Chang and Ghamisi(2023)]{chg2cap}
Shizhen Chang and Pedram Ghamisi.
\newblock Changes to captions: An attentive network for remote sensing change captioning.
\newblock \emph{TIP}, 2023.

\bibitem[Chiang et~al.(2023)Chiang, Li, Lin, Sheng, Wu, Zhang, Zheng, Zhuang, Zhuang, Gonzalez, et~al.]{chiang2023vicuna}
Wei-Lin Chiang, Zhuohan Li, Zi Lin, Ying Sheng, Zhanghao Wu, Hao Zhang, Lianmin Zheng, Siyuan Zhuang, Yonghao Zhuang, Joseph~E Gonzalez, et~al.
\newblock Vicuna: An open-source chatbot impressing gpt-4 with 90\%* chatgpt quality.
\newblock \emph{Technical Report}, 2023.

\bibitem[Chung et~al.(2019)Chung, Kraska, Polyzotis, Tae, and Whang]{chung2019automated}
Yeounoh Chung, Tim Kraska, Neoklis Polyzotis, Ki~Hyun Tae, and Steven~Euijong Whang.
\newblock Automated data slicing for model validation:a big data - ai integration approach.
\newblock In \emph{ICDE}, 2019.

\bibitem[Deng et~al.(2009)Deng, Dong, Socher, Li, Li, and Fei-Fei]{deng2009imagenet}
Jia Deng, Wei Dong, Richard Socher, Li-Jia Li, Kai Li, and Li Fei-Fei.
\newblock Imagenet: A large-scale hierarchical image database.
\newblock In \emph{CVPR}, 2009.

\bibitem[d'Eon et~al.(2021)d'Eon, d'Eon, Wright, and Leyton-Brown]{deon2021spotlight}
Greg d'Eon, Jason d'Eon, James~R. Wright, and Kevin Leyton-Brown.
\newblock The spotlight: A general method for discovering systematic errors in deep learning models.
\newblock In \emph{FAccT}, 2021.

\bibitem[Digital(2018)]{Digital-2018}
by:~Power Digital.
\newblock Instagram algorithm change history, 2018.

\bibitem[Doersch et~al.(2012)Doersch, Singh, Gupta, Sivic, and Efros]{doersch2012what}
Carl Doersch, Saurabh Singh, Abhinav Gupta, Josef Sivic, and Alexei~A. Efros.
\newblock What makes paris look like paris?
\newblock In \emph{SIGGRAPH}, 2012.

\bibitem[Dubois et~al.(2023)Dubois, Li, Taori, Zhang, Gulrajani, Ba, Guestrin, Liang, and Hashimoto]{dubois2023alpacafarm}
Yann Dubois, Xuechen Li, Rohan Taori, Tianyi Zhang, Ishaan Gulrajani, Jimmy Ba, Carlos Guestrin, Percy Liang, and Tatsunori~B Hashimoto.
\newblock Alpacafarm: A simulation framework for methods that learn from human feedback.
\newblock \emph{arXiv preprint arXiv:2305.14387}, 2023.

\bibitem[Engstrom et~al.(2020)Engstrom, Ilyas, Santurkar, Tsipras, Steinhardt, and Madry]{engstrom2020identifying}
Logan Engstrom, Andrew Ilyas, Shibani Santurkar, Dimitris Tsipras, Jacob Steinhardt, and Aleksander Madry.
\newblock Identifying statistical bias in dataset replication.
\newblock In \emph{ICML}, 2020.

\bibitem[Eyuboglu et~al.(2022)Eyuboglu, Varma, Saab, Delbrouck, Lee-Messer, Dunnmon, Zou, and Re]{eyuboglu2021domino}
Sabri Eyuboglu, Maya Varma, Khaled~Kamal Saab, Jean-Benoit Delbrouck, Christopher Lee-Messer, Jared Dunnmon, James Zou, and Christopher Re.
\newblock Domino: Discovering systematic errors with cross-modal embeddings.
\newblock In \emph{ICLR}, 2022.

\bibitem[Goh et~al.(2021)Goh, Cammarata, Voss, Carter, Petrov, Schubert, Radford, and Olah]{goh2021multimodal}
Gabriel Goh, Nick Cammarata, Chelsea Voss, Shan Carter, Michael Petrov, Ludwig Schubert, Alec Radford, and Chris Olah.
\newblock Multimodal neurons in artificial neural networks.
\newblock \emph{Distill}, 2021.

\bibitem[He et~al.(2016)He, Zhang, Ren, and Sun]{he2016deep}
Kaiming He, Xiangyu Zhang, Shaoqing Ren, and Jian Sun.
\newblock Deep residual learning for image recognition.
\newblock In \emph{CVPR}, 2016.

\bibitem[Hendrycks et~al.(2021)Hendrycks, Basart, Mu, Kadavath, Wang, Dorundo, Desai, Zhu, Parajuli, Guo, Song, Steinhardt, and Gilmer]{hendrycks2021many}
Dan Hendrycks, Steven Basart, Norman Mu, Saurav Kadavath, Frank Wang, Evan Dorundo, Rahul Desai, Tyler Zhu, Samyak Parajuli, Mike Guo, Dawn Song, Jacob Steinhardt, and Justin Gilmer.
\newblock The many faces of robustness: A critical analysis of out-of-distribution generalization.
\newblock In \emph{ICCV}, 2021.

\bibitem[Hernandez et~al.(2021)Hernandez, Schwettmann, Bau, Bagashvili, Torralba, and Andreas]{hernandez2021natural}
Evan Hernandez, Sarah Schwettmann, David Bau, Teona Bagashvili, Antonio Torralba, and Jacob Andreas.
\newblock Natural language descriptions of deep visual features.
\newblock In \emph{ICLR}, 2021.

\bibitem[Isola et~al.(2011{\natexlab{a}})Isola, Parikh, Torralba, and Oliva]{IsolaParikhTorralbaOliva2011}
Phillip Isola, Devi Parikh, Antonio Torralba, and Aude Oliva.
\newblock Understanding the intrinsic memorability of images.
\newblock In \emph{NeurIPS}, 2011{\natexlab{a}}.

\bibitem[Isola et~al.(2011{\natexlab{b}})Isola, Xiao, Torralba, and Oliva]{isola2011makes}
Phillip Isola, Jianxiong Xiao, Antonio Torralba, and Aude Oliva.
\newblock What makes an image memorable?
\newblock In \emph{CVPR}, 2011{\natexlab{b}}.

\bibitem[Jain et~al.(2023)Jain, Lawrence, Moitra, and Madry]{jain2023distilling}
Saachi Jain, Hannah Lawrence, Ankur Moitra, and Aleksander Madry.
\newblock Distilling model failures as directions in latent space.
\newblock In \emph{ICLR}, 2023.

\bibitem[Kim et~al.(2021)Kim, Kim, Lee, Park, and Kim]{kim2021viewpoint}
Hoeseong Kim, Jongseok Kim, Hyungseok Lee, Hyunsung Park, and Gunhee Kim.
\newblock Viewpoint-agnostic change captioning with cycle consistency.
\newblock In \emph{ICCV}, 2021.

\bibitem[Koh et~al.(2021)Koh, Sagawa, Marklund, Xie, Zhang, Balsubramani, Hu, Yasunaga, Phillips, Gao, et~al.]{koh2021wilds}
Pang~Wei Koh, Shiori Sagawa, Henrik Marklund, Sang~Michael Xie, Marvin Zhang, Akshay Balsubramani, Weihua Hu, Michihiro Yasunaga, Richard~Lanas Phillips, Irena Gao, et~al.
\newblock Wilds: A benchmark of in-the-wild distribution shifts.
\newblock In \emph{ICML}, 2021.

\bibitem[Li et~al.(2023{\natexlab{a}})Li, Zhang, Chen, Wang, Pu, Yang, Li, and Liu]{li2023mimicit}
Bo Li, Yuanhan Zhang, Liangyu Chen, Jinghao Wang, Fanyi Pu, Jingkang Yang, Chunyuan Li, and Ziwei Liu.
\newblock Mimic-it: Multi-modal in-context instruction tuning.
\newblock \emph{arXiv preprint arXiv:2306.05425}, 2023{\natexlab{a}}.

\bibitem[Li et~al.(2023{\natexlab{b}})Li, Zhang, Chen, Wang, Yang, and Liu]{li2023otter}
Bo Li, Yuanhan Zhang, Liangyu Chen, Jinghao Wang, Jingkang Yang, and Ziwei Liu.
\newblock Otter: A multi-modal model with in-context instruction tuning.
\newblock \emph{arXiv preprint arXiv:2305.03726}, 2023{\natexlab{b}}.

\bibitem[Li et~al.(2023{\natexlab{c}})Li, Li, Savarese, and Hoi]{li2023blip}
Junnan Li, Dongxu Li, Silvio Savarese, and Steven Hoi.
\newblock Blip-2: Bootstrapping language-image pre-training with frozen image encoders and large language models.
\newblock \emph{arXiv preprint arXiv:2301.12597}, 2023{\natexlab{c}}.

\bibitem[Liang and Zou(2022)]{liang2021metashift}
Weixin Liang and James Zou.
\newblock Metashift: A dataset of datasets for evaluating contextual distribution shifts and training conflicts.
\newblock In \emph{ICLR}, 2022.

\bibitem[Liu et~al.(2023{\natexlab{a}})Liu, Li, Wu, and Lee]{liu2023visual}
Haotian Liu, Chunyuan Li, Qingyang Wu, and Yong~Jae Lee.
\newblock Visual instruction tuning.
\newblock \emph{arXiv preprint arXiv:2304.08485}, 2023{\natexlab{a}}.

\bibitem[Liu et~al.(2023{\natexlab{b}})Liu, Lin, Hewitt, Paranjape, Bevilacqua, Petroni, and Liang]{lost_in_the_middle}
Nelson~F. Liu, Kevin Lin, John Hewitt, Ashwin Paranjape, Michele Bevilacqua, Fabio Petroni, and Percy Liang.
\newblock Lost in the middle: How language models use long contexts.
\newblock \emph{arXiv preprint arXiv:2307.03172}, 2023{\natexlab{b}}.

\bibitem[Lundberg and Lee(2017)]{lundberg2017unified}
Scott~M Lundberg and Su-In Lee.
\newblock A unified approach to interpreting model predictions.
\newblock In \emph{NeurIPS}, 2017.

\bibitem[Manovich(2012)]{Manovich2012}
Lev Manovich.
\newblock \emph{How to Compare One Million Images?}, pages 249--278.
\newblock Palgrave Macmillan UK, London, 2012.

\bibitem[Miller et~al.(2021)Miller, Taori, Raghunathan, Sagawa, Koh, Shankar, Liang, Carmon, and Schmidt]{miller2021accuracy}
John~P Miller, Rohan Taori, Aditi Raghunathan, Shiori Sagawa, Pang~Wei Koh, Vaishaal Shankar, Percy Liang, Yair Carmon, and Ludwig Schmidt.
\newblock Accuracy on the line: on the strong correlation between out-of-distribution and in-distribution generalization.
\newblock In \emph{ICML}, 2021.

\bibitem[OpenAI(2023)]{openai2023gpt4}
OpenAI.
\newblock Gpt-4 technical report.
\newblock \emph{arXiv preprint arXiv:2303.08774}, 2023.

\bibitem[Park et~al.(2019)Park, Darrell, and Rohrbach]{robust_change_captioning}
Dong~Huk Park, Trevor Darrell, and Anna Rohrbach.
\newblock Robust change captioning.
\newblock In \emph{ICCV}, 2019.

\bibitem[Park et~al.(2023)Park, Georgiev, Ilyas, Leclerc, and Madry]{park2023trak}
Sung~Min Park, Kristian Georgiev, Andrew Ilyas, Guillaume Leclerc, and Aleksander Madry.
\newblock Trak: Attributing model behavior at scale.
\newblock In \emph{ICML}, 2023.

\bibitem[Quinonero-Candela et~al.(2008)Quinonero-Candela, Sugiyama, Schwaighofer, and Lawrence]{quinonero2008dataset}
Joaquin Quinonero-Candela, Masashi Sugiyama, Anton Schwaighofer, and Neil~D Lawrence.
\newblock \emph{Dataset shift in machine learning}.
\newblock Mit Press, 2008.

\bibitem[Radford et~al.(2021)Radford, Kim, Hallacy, Ramesh, Goh, Agarwal, Sastry, Askell, Mishkin, Clark, et~al.]{radford2021learning}
Alec Radford, Jong~Wook Kim, Chris Hallacy, Aditya Ramesh, Gabriel Goh, Sandhini Agarwal, Girish Sastry, Amanda Askell, Pamela Mishkin, Jack Clark, et~al.
\newblock Learning transferable visual models from natural language supervision.
\newblock In \emph{ICML}, 2021.

\bibitem[Recht et~al.(2019)Recht, Roelofs, Schmidt, and Shankar]{recht2019imagenet}
Benjamin Recht, Rebecca Roelofs, Ludwig Schmidt, and Vaishaal Shankar.
\newblock Do imagenet classifiers generalize to imagenet?
\newblock In \emph{ICML}, 2019.

\bibitem[Redmon et~al.(2016)Redmon, Divvala, Girshick, and Farhadi]{redmon2016you}
Joseph Redmon, Santosh Divvala, Ross Girshick, and Ali Farhadi.
\newblock You only look once: Unified, real-time object detection.
\newblock In \emph{CVPR}, 2016.

\bibitem[Ribeiro et~al.(2016)Ribeiro, Singh, and Guestrin]{ribeiro2016should}
Marco~Tulio Ribeiro, Sameer Singh, and Carlos Guestrin.
\newblock "why should i trust you?" explaining the predictions of any classifier.
\newblock In \emph{KDD}, 2016.

\bibitem[Rombach et~al.(2022)Rombach, Blattmann, Lorenz, Esser, and Ommer]{rombach2022high}
Robin Rombach, Andreas Blattmann, Dominik Lorenz, Patrick Esser, and Bj{\"o}rn Ommer.
\newblock High-resolution image synthesis with latent diffusion models.
\newblock In \emph{CVPR}, 2022.

\bibitem[Shah et~al.(2023)Shah, Park, Ilyas, and Madry]{shah2023modeldiff}
Harshay Shah, Sung~Min Park, Andrew Ilyas, and Aleksander Madry.
\newblock Modeldiff: A framework for comparing learning algorithms.
\newblock In \emph{ICML}, 2023.

\bibitem[Shi et~al.(2023)Shi, Chen, Misra, Scales, Dohan, Chi, Schärli, and Zhou]{shi2023large}
Freda Shi, Xinyun Chen, Kanishka Misra, Nathan Scales, David Dohan, Ed Chi, Nathanael Schärli, and Denny Zhou.
\newblock Large language models can be easily distracted by irrelevant context.
\newblock \emph{arXiv preprint arXiv:2302.00093}, 2023.

\bibitem[Torralba and Efros(2011)]{unbiased_look}
Antonio Torralba and Alexei Efros.
\newblock Unbiased look at dataset bias.
\newblock In \emph{CVPR}, 2011.

\bibitem[van Noord(2023)]{dataset_comparison}
Nanne van Noord.
\newblock Prototype-based dataset comparison.
\newblock In \emph{ICCV}, 2023.

\bibitem[Vendrow et~al.(2023)Vendrow, Jain, Engstrom, and Madry]{vendrow2023dataset}
Joshua Vendrow, Saachi Jain, Logan Engstrom, and Aleksander Madry.
\newblock Dataset interfaces: Diagnosing model failures using controllable counterfactual generation.
\newblock \emph{arXiv preprint arXiv:2302.07865}, 2023.

\bibitem[Wang et~al.(2020)Wang, Narayanan, and Russakovsky]{revisetool_eccv}
Angelina Wang, Arvind Narayanan, and Olga Russakovsky.
\newblock {REVISE}: A tool for measuring and mitigating bias in visual datasets.
\newblock In \emph{ECCV}, 2020.

\bibitem[Wang et~al.(2022)Wang, Montoya, Munechika, Yang, Hoover, and Chau]{wangDiffusionDBLargescalePrompt2022}
Zijie~J. Wang, Evan Montoya, David Munechika, Haoyang Yang, Benjamin Hoover, and Duen~Horng Chau.
\newblock {{DiffusionDB}}: {{A}} large-scale prompt gallery dataset for text-to-image generative models.
\newblock \emph{arXiv preprint arXiv:2210.14896}, 2022.

\bibitem[Yao et~al.(2022)Yao, Wang, and Jin]{yao2022imagediff}
Linli Yao, Weiying Wang, and Qin Jin.
\newblock Image difference captioning with pre-training and contrastive learning.
\newblock In \emph{AAAI}, 2022.

\bibitem[Yu et~al.(2022)Yu, Xu, Koh, Luong, Baid, Wang, Vasudevan, Ku, Yang, Ayan, et~al.]{parti}
Jiahui Yu, Yuanzhong Xu, Jing~Yu Koh, Thang Luong, Gunjan Baid, Zirui Wang, Vijay Vasudevan, Alexander Ku, Yinfei Yang, Burcu~Karagol Ayan, et~al.
\newblock Scaling autoregressive models for content-rich text-to-image generation.
\newblock \emph{TMLR}, 2022.

\bibitem[Zheng et~al.(2023)Zheng, Chiang, Sheng, Zhuang, Wu, Zhuang, Lin, Li, Li, Xing, Zhang, Gonzalez, and Stoica]{zheng2023judging}
Lianmin Zheng, Wei-Lin Chiang, Ying Sheng, Siyuan Zhuang, Zhanghao Wu, Yonghao Zhuang, Zi Lin, Zhuohan Li, Dacheng Li, Eric.~P Xing, Hao Zhang, Joseph~E. Gonzalez, and Ion Stoica.
\newblock Judging llm-as-a-judge with mt-bench and chatbot arena.
\newblock In \emph{NeurIPS Datasets and Benchmarks}, 2023.

\bibitem[Zhong et~al.(2022)Zhong, Snell, Klein, and Steinhardt]{zhongd3}
Ruiqi Zhong, Charlie Snell, Dan Klein, and Jacob Steinhardt.
\newblock Describing differences between text distributions with natural language.
\newblock In \emph{ICML}, 2022.

\bibitem[Zhong et~al.(2023)Zhong, Zhang, Li, Ahn, Klein, and Steinhardt]{zhongd5}
Ruiqi Zhong, Peter Zhang, Steve Li, Jinwoo Ahn, Dan Klein, and Jacob Steinhardt.
\newblock Goal driven discovery of distributional differences via language descriptions.
\newblock \emph{arXiv preprint arXiv:2302.14233}, 2023.

\bibitem[Zhou et~al.(2016)Zhou, Khosla, Lapedriza, Oliva, and Torralba]{zhou2016cvpr}
Bolei Zhou, Aditya Khosla, Agata Lapedriza, Aude Oliva, and Antonio Torralba.
\newblock Learning deep features for discriminative localization.
\newblock In \emph{CVPR}, 2016.

\bibitem[Zhu et~al.(2014)Zhu, Lee, and Efros]{zhu2014averageExplorer}
Jun-Yan Zhu, Yong~Jae Lee, and Alexei~A Efros.
\newblock Averageexplorer: Interactive exploration and alignment of visual data collections.
\newblock In \emph{SIGGRAPH}, 2014.

\bibitem[Zhu et~al.(2022)Zhu, Liang, and Zou]{zhu2022gsclip}
Zhiying Zhu, Weixin Liang, and James Zou.
\newblock Gsclip: A framework for explaining distribution shifts in natural language.
\newblock In \emph{ICML DataPerf Workshop}, 2022.

\end{thebibliography}
}

\appendix
\maketitlesupplementary

\section*{Acknowledgements} 

We thank all the reviewers for their constructive feedback. We thank James Zou, Weixin Liang, Jeff Z. HaoChen, Jen Weng, Zeyu Wang, Jackson Wang, Elaine Sui, Ruocheng Wang for providing valuable feedback to this project. We also thank Dan Klein for providing feedback on the abstract and intro as well as Esau Hutcherson and Yannis Siglidis for running preliminary experiments on VisDiffBench and the LaMem dataset. Lastly, we thank Alexei Efros for proposing several dozen applications, providing relevant related works, and for grudgingly acknowledging that the task of  \task{} is ``cool, even though it has language''. This work was supported in part by the NSF CISE Expeditions Award (CCF-1730628). Trevor Darrell and Lisa Dunlap were supported by DoD and/or BAIR Industrial funds. Serena Yeung-Levy is a Chan Zuckerberg Biohub --- San Francisco Investigator.

\section*{Reproducibility Statement}

We provide code implementations of \method{} at \url{https://github.com/Understanding-Visual-Datasets/VisDiff}. 
We also provide \benchmark{} at \url{https://drive.google.com/file/d/1vghFd0rB5UTBaeR5rdxhJe3s7OOdRtkY}.
The implementations and datasets will enable researchers to reproduce all the experiments described in the paper as well as run their own analyses on additional datasets.

\section*{Ethics Statement}

In this work, we introduce \method{}, a novel method designed to discern subtle differences between two sets of images. \method{} represents not just a technical advance in the analysis of image sets, but also serves as a useful tool to promote fairness, diversity, and scientific discovery in AI and data science.
First, \method{} has the potential to \emph{uncover biases} in datasets. For instance, comparing image sets of workers from diverse demographic groups, such as men and women, can reveal and quantify career stereotypes associated with each group. This capability is pivotal in addressing and mitigating biases present in datasets.
Furthermore, \method{} holds substantial promise for \emph{scientific discovery}. By comparing image sets in various scientific domains, such as cellular images from patients and healthy individuals, \method{} can unveil novel insights into the disease impacts on cellular structures, thereby driving forward critical advancements in medical research.
However, \method{} is meant to be an \emph{assistive tool} and should be applied with humans in the loop. The users are responsible for interpreting the results properly and avoiding misinformation.
In summary, \method{} emerges as a crucial tool for ethical AI considerations, fostering fairness and catalyzing scientific progress.

\section*{Table of Contents}

In this supplementary material, we provide additional details of datasets, methods, results, and applications.

\begin{itemize}
    \item In~\autoref{sec:supp_sec3}, we provide examples of our benchmark \benchmark{} prompts to generate and evaluate this benchmark, human-generated labels for \benchmark{}, and Other \benchmark{} evaluation metrics.
    \item In~\autoref{sec:supp_sec4}, we provide additional details of each proposer and ranker and compare different ranking metrics.
    \item In~\autoref{sec:supp_sec5}, we ablate various design choices of our algorithm \method{}.
    \item In~\autoref{sec:supp_sec6}, we provide supplementary evidence of findings for each application.
    \item In~\autoref{sec:supp_limitations}, we explain more failure cases and limitations of \method{}.
\end{itemize}

\section{Supplementary Section 3}
\label{sec:supp_sec3}

In this section, we provide additional details of Section 3 in the main paper.

\subsection{Paired Sentences for \benchmark{}}
\label{sec:visdiff_paried_sentences}

\benchmark{} contains five subsets: \ourbenchmark{}-Easy, \ourbenchmark{}-Medium, \ourbenchmark{}-Hard, ImageNetR, and ImageNet$^*$. We provide all the paired sentences of \ourbenchmark{} in~\autoref{supp_tab:paired_image_sets}. For ImageNetR, \seta{} is one of the ``art'', ``cartoon'', ``deviantart'', ``embroidery'', ``graffiti'', ``graphic'', ``origami'', ``painting'', ``sculpture'', ``sketch'', ``sticker'', ``tattoo'', ``toy'', ``videogame'', and \setb{} is ``imagenet''. For ImageNet$^*$, \seta{} is one of the ``in the forest'', ``green'', ``red'', ``pencil sketch'', ``oil painting'', ``orange'', ``on the rocks'', ``in bright sunlight'', ``person and a'', ``in the beach'', ``studio lighting'', ``in the water'', ``at dusk'', ``in the rain'', ``in the grass'', ``yellow'', ``blue'', ``and a flower'', ``on the road'', ``at night'', ``embroidery'', ``in the fog'', ``in the snow'', and \setb{} is ``base''.

\begin{table*}[!tb]
\centering
\tiny
\begin{tabular}{p{0.14\linewidth}p{0.14\linewidth}|p{0.14\linewidth}p{0.14\linewidth}|p{0.14\linewidth}p{0.14\linewidth}}
\toprule
\multicolumn{2}{c|}{\textbf{Easy (50 Paired Sets)}} & \multicolumn{2}{c|}{\textbf{Medium (50 Paired Sets)}} & \multicolumn{2}{c}{\textbf{Hard (50 Paired Sets)}} \\
\textbf{Set A} & \textbf{Set B} & \textbf{Set A} & \textbf{Set B} & \textbf{Set A} & \textbf{Set B} \\
\midrule
Dogs playing in a park & Cats playing in a park & SUVs on a road & Sedans on a road & Sunrise over Santorini, Greece & Sunset over Santorini, Greece \\
Children playing soccer & Children swimming in a pool & Wooden chairs in a room & Plastic chairs in a room & People practicing yoga in a mountainous setting & People meditating in a mountainous setting \\
Snow-covered mountains & Desert sand dunes & Golden retriever dogs playing & Labrador dogs playing & Fresh sushi with salmon topping & Fresh sushi with tuna topping \\
Butterflies on flowers & Bees on flowers & Green apples in a basket & Red apples in a basket & Lush vineyards in spring & Lush vineyards in early autumn \\
People shopping in a mall & People dining in a restaurant & Leather shoes on display & Canvas shoes on display & Men wearing Rolex watches & Men wearing Omega watches \\
Elephants in the savannah & Giraffes in the savannah & Freshwater fish in an aquarium & Saltwater fish in an aquarium & Cupcakes topped with buttercream & Cupcakes topped with fondant \\
Birds flying in the sky & Airplanes flying in the sky & Steel bridges over a river & Wooden bridges over a river & People playing chess outdoors & People playing checkers outdoors \\
Boats in a marina & Cars in a parking lot & Mountain bikes on a trail & Road bikes on a road & Hand-painted porcelain plates & Hand-painted ceramic plates \\
Tulips in a garden & Roses in a garden & Ceramic mugs on a shelf & Glass mugs on a shelf & Cyclists in a time-trial race & Cyclists in a mountain stage race \\
People skiing on a slope & People snowboarding on a slope & People playing electric guitars & People playing acoustic guitars & Gardens with Japanese cherry blossoms & Gardens with Japanese maples \\
Fish in an aquarium & Turtles in an aquarium & Laptop computers on a desk & Desktop computers on a desk & People wearing traditional Korean hanboks & People wearing traditional Japanese kimonos \\
Books on a shelf & Plants on a shelf & Hardcover books on a table & Paperback books on a table & Alpine lakes in summer & Alpine lakes in early spring \\
Grapes in a bowl & Apples in a bowl & Digital clocks on a wall & Analog clocks on a wall & Merlot wine in a glass & Cabernet Sauvignon wine in a glass \\
Motorcycles on a street & Bicycles on a street & Children playing with toy cars & Children playing with toy trains & Football players in defensive formation & Football players in offensive formation \\
Cows grazing in a field & Sheep grazing in a field & White roses in a vase & Pink roses in a vase & Classic novels from the 19th century & Modern novels from the 21st century \\
Babies in cribs & Babies in strollers & Electric stoves in a kitchen & Gas stoves in a kitchen & Orchestras playing Baroque music & Orchestras playing Classical music \\
Hot air balloons in the air & Kites in the air & Leather jackets on hangers & Denim jackets on hangers & Men in British army uniforms from WWI & Men in British army uniforms from WWII \\
Penguins in the snow & Seals in the snow & People eating with chopsticks & People eating with forks & Sculptures from the Renaissance era & Sculptures from the Hellenistic era \\
Lions in a jungle & Monkeys in a jungle & Pearl necklaces on display & Gold necklaces on display & People preparing macarons & People preparing meringues \\
Watches on a display & Rings on a display & Mushrooms in a forest & Ferns in a forest & Female ballet dancers in pointe shoes & Female ballet dancers in ballet slippers \\
Pizzas in a box & Donuts in a box & Stainless steel kettles in a store & Plastic kettles in a store & Dishes from Northern Italian cuisine & Dishes from Southern Italian cuisine \\
Bricks on a wall & Tiles on a wall & Porcelain vases on a shelf & Metal vases on a shelf & Classic rock bands performing & Alternative rock bands performing \\
Pianos in a room & Guitars in a room & Vintage cars on a road & Modern cars on a road & Historical films set in Medieval Europe & Historical films set in Ancient Rome \\
Trains on tracks & Buses on roads & Handmade quilts on a bed & Factory-made blankets on a bed & Bonsai trees shaped in cascade style & Bonsai trees shaped in informal upright style \\
Pots on a stove & Plates on a table & Shiny silk dresses on mannequins & Matte cotton dresses on mannequins & Lace wedding dresses & Satin wedding dresses \\
Stars in the night sky & Clouds in the day sky & Mechanical pencils on a desk & Ballpoint pens on a desk & Birds with iridescent plumage & Birds with matte plumage \\
Sunflowers in a field & Wheat in a field & Ginger cats lying down & Tabby cats lying down & Women wearing matte lipstick & Women wearing glossy lipstick \\
Dolls on a shelf & Teddy bears on a shelf & People riding racing horses & People riding dressage horses & Cities with Gothic architecture & Cities with Modernist architecture \\
Pine trees in a forest & Oak trees in a forest & Steel water bottles on a table & Glass water bottles on a table & Poems written in free verse & Poems written in sonnet form \\
Men playing basketball & Women playing volleyball & Men wearing leather gloves & Men wearing wool gloves & Acoustic guitars being played & Classical guitars being played \\
Ice cream in a cone & Juice in a glass & Rubber ducks in a tub & Plastic boats in a tub & Books with hardcover binding & Books with leather-bound covers \\
Dancers on a stage & Singers on a stage & Porcelain tea cups on a tray & Glass tea cups on a tray & Portraits painted in cubist style & Portraits painted in impressionist style \\
Rainbows in the sky & Lightning in the sky & Sparrows on a tree & Canaries on a tree & Residential buildings in Art Deco style & Residential buildings in Brutalist style \\
Towers in a city & Houses in a suburb & Shiny metallic cars & Matte finish cars & Male professional swimmers in freestyle race & Male professional swimmers in butterfly race \\
Frogs by a pond & Ducks by a pond & Stuffed teddy bears on a bed & Stuffed bunny rabbits on a bed & Basketball players attempting free throws & Basketball players attempting slam dunks \\
Football players on a field & Rugby players on a field & Round dinner plates on a table & Square dinner plates on a table & Cakes decorated with marzipan & Cakes decorated with buttercream roses \\
Pillows on a bed & Blankets on a bed & Butter on a slice of bread & Jam on a slice of bread & People practicing the Sun Salutation in yoga & People practicing the Tree Pose in yoga \\
Deer in a forest & Rabbits in a forest & Bengal cat in sitting posture & Siamese cat in sitting posture & Men wearing suits & Men wearing tuxedos \\
Tea in a cup & Coffee in a cup & Violinists playing in a quartet & Cellists playing in a quartet & Butterflies with spotted wings & Butterflies with striped wings \\
Children on a slide & Children on a swing & Gothic cathedrals in Europe & Baroque churches in Europe & Oak trees in summer & Oak trees in autumn \\
Kangaroos in a desert & Camels in a desert & People dancing tango & People dancing waltz & Tennis shoes on a rack & Running shoes on a rack \\
Tomatoes in a basket & Eggs in a basket & Abstract oil paintings with warm colors & Abstract oil paintings with cool colors & People playing classical violin & People playing fiddle \\
People in an elevator & People on an escalator & Candies made from dark chocolate & Candies made from milk chocolate & Men wearing fedoras & Men wearing baseball caps \\
Sandcastles on a beach & Umbrellas on a beach & Rivers in tropical rainforests & Rivers in alpine meadows & Passenger planes in the sky & Cargo planes in the sky \\
Mice in a barn & Horses in a barn & Cars from the 1960s & Cars from the 1980s & Women wearing ankle boots & Women wearing knee-high boots \\
Chocolates in a box & Candies in a jar & Seascapes during a storm & Seascapes during a calm day & Diesel trucks on a highway & Electric trucks on a highway \\
Zebra crossings on a street & Traffic lights on a street & Fruits arranged in a still life setting & Flowers arranged in a still life setting & Children reading comic books & Children reading fairy tales \\
Bridges over a river & Boats on a river & Dishes from Thai cuisine & Dishes from Vietnamese cuisine & Men wearing round glasses & Men wearing square glasses \\
Oranges on a tree & Bird nests on a tree & Wild horses in American plains & Wild zebras in African savannahs & Vinyl records in a store & CDs in a store \\
Lanterns in a festival & Fireworks in a festival & Classic movies in black and white & Classic movies in Technicolor & Bonsai trees in pots & Cacti in pots \\
\bottomrule
\end{tabular}
\caption{Paired sentences for \ourbenchmark{}. Easy, medium, and hard examples are shown in the left, middle, and right.}
\label{supp_tab:paired_image_sets}
\end{table*}

\subsection{Examples for \benchmark{}} 

\begin{figure*}[!tb]
     \centering
         \includegraphics[width=0.92\textwidth]{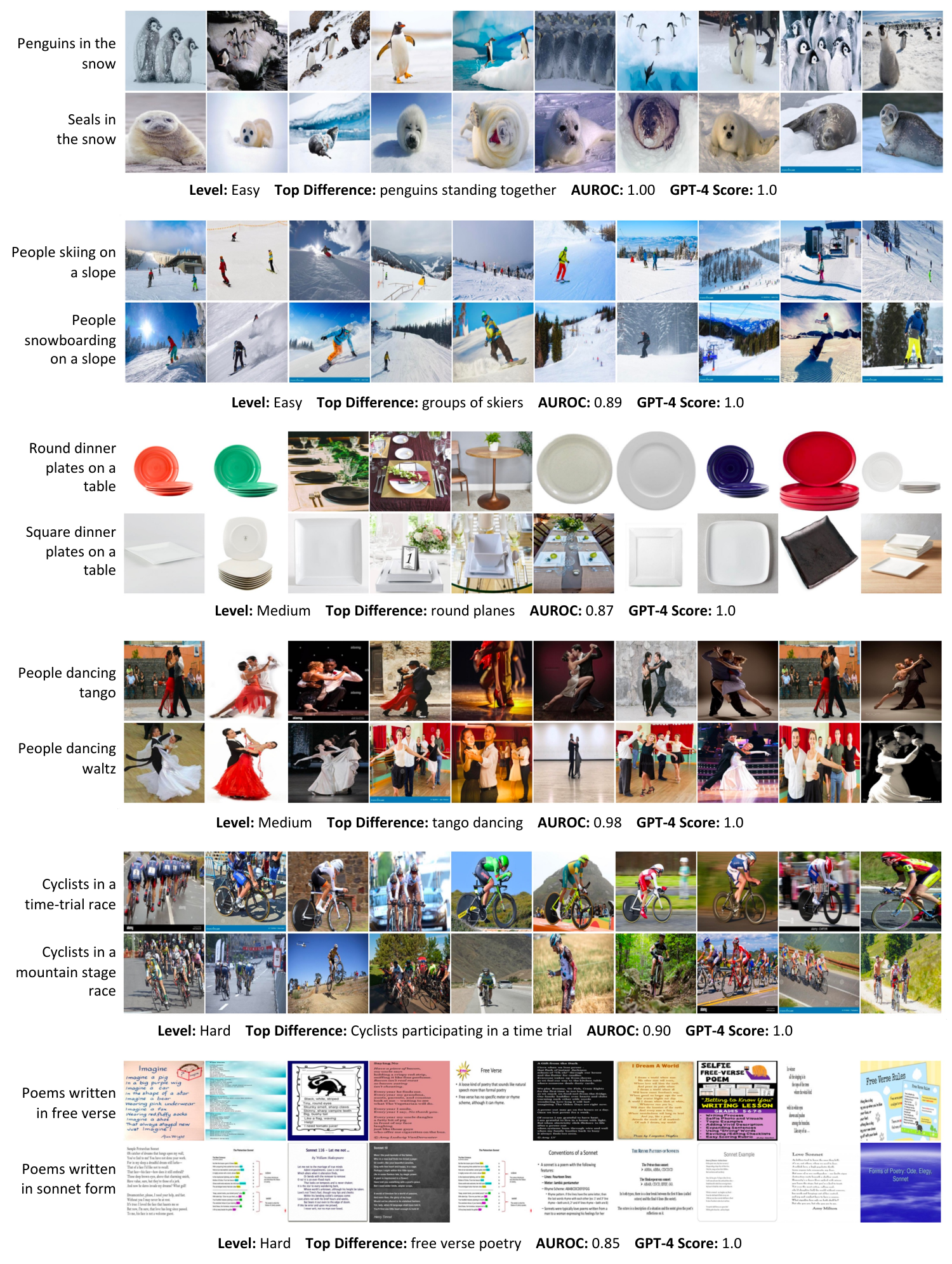}
        \caption{\ourbenchmark{} examples where \method{} succeeds. We show the ground-truth difference, top difference predicted by \method{}, AUROC score output by the ranker, and evaluation of the predicted difference by GPT-4.}
        \label{supp_fig:successful_visdiff_examples}
\end{figure*}

We provide 4 examples for \ourbenchmark{}-Easy, \ourbenchmark{}-Medium, \ourbenchmark{}-Hard, respectively, in~\autoref{supp_fig:successful_visdiff_examples} and~\autoref{supp_fig:failed_visdiff_examples}. For ImageNetR and ImageNet$^*$, we refer readers to the original papers~\cite{hendrycks2021many,vendrow2023dataset}.

\subsection{Prompts for \benchmark{} Generation}
\label{sec:visdiff_generation_prompts}

We provide the GPT-4 prompt we used to generate paired sentences for \ourbenchmark{} in~\autoref{supp_fig:supp_sec3_prompt} (top). 

\subsection{Prompts for \benchmark{} Evaluation}
\label{sec:visdiff_eval_prompts}

\begin{figure*}[!tb]
    \centering
    \includegraphics[width=\textwidth]{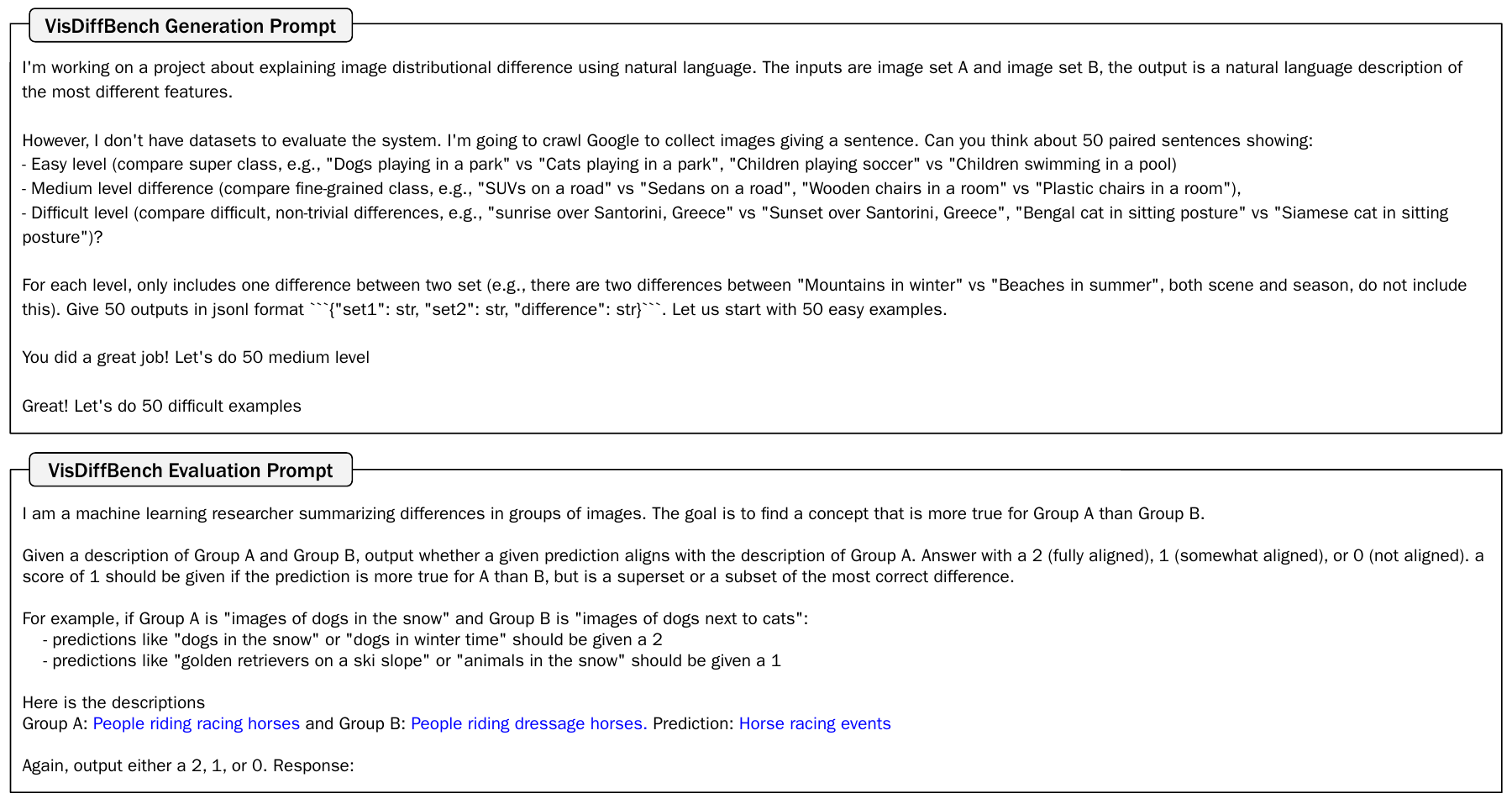}
    \caption{Prompt used to generate paired sentences for \benchmark{} (top) and evaluate \benchmark{} (bottom). Input-dependent texts are colored in blue.}
    \label{supp_fig:supp_sec3_prompt}
\end{figure*}

We provide the GPT-4 prompt we used to evaluate the generated difference description against the ground-truth difference description in~\autoref{supp_fig:supp_sec3_prompt} (bottom).

\subsection{Human-generated Differences for \benchmark{}}

To increase the quality of the dataset, we have collected human-generated differences between the sets in \benchmark{}. We have conducted two types of human annotations: \textbf{(1)} propose the differences by humans; \textbf{(2)} validate the differences by humans. Averaged across \textbf{3} annotations for each of \textbf{187} sets, we find that annotators agreed that \textbf{96\%} of our labels are correct differences, \textbf{93\%} are the best description to differentiate the set, and \textbf{76\%} are the same as a difference the annotator has written. The last statistic is indicative of human performance on this challenging task. Since this task has some difficult cases, for instance, when set A is ``cities with Gothic architecture'' and set B is ``cities with Modernist architecture'', we see models outperform humans on some cases. 

In the first part of annotation, annotators are given the link to the images from $\mathcal{D_A}$ and $\mathcal{D_B}$ and asked to propose up to 5 differences (usually 1-2). In the second part, the annotators are given our VisDiffBench ground-truth descriptions and asked (1) is the provided difference correct (2) would you consider this the best description of the difference between the sets (3) is this consistent with any of your descriptions and (4) which description is it most consistent with. We gave each annotator a tutorial on the task with 3 examples, checking their first few descriptions were in the correct format. In the end we collected 3 annotations per image set in VisDiffBench. The inter-annotator agreement is 93\%, 87\%, 75\% for questions 1-3. We have released these human labels along with our original labels in our code base.

\subsection{Other \benchmark{} Evaluation Metrics}

We chose GPT-4 as our evaluation metric because evaluating the proposed difference requires a human-level understanding of the semantics in a short description.
\autoref{tab:automatic_metric} reports the correlation between common captioning metrics and human annotators in VisDiffBench, which shows that the GPT-4 evaluation has much higher consensus with humans and is the only reliable metric. 
However, due to the limitation of GPT-4 being closed-source and constantly changing, we highlighted the exact GPT version we used (\textit{gpt-4-0613}) and released the outputs of our experiments to maximize reproducibility.

\begin{table}[!tb]
    \scriptsize
    \centering
    \setlength{\tabcolsep}{3.5pt}
    \resizebox{\linewidth}{!}{ 
    \begin{tabular}{l|ccc|cccc}
        \toprule
        \textbf{Metric} & B4 & R1 & RL & BS & V1.5 & G3.5 & G4 \\
        \midrule
        \textbf{Pearson} & 0.140 & 0.492 & 0.497 & 0.272 & 0.594 & 0.623 & \textbf{0.800} \\
        \bottomrule
    \end{tabular}
    }
    \caption{\textbf{Correlation of automated metric with humans.} Model-free metrics include B4 (BLEU-4), R1 (ROUGE-1), RL (ROUGE-L). Model-based metrics include BS (BERTScore), V1.5 (Vicuna-1.5-13B), G3.5 (gpt-3.5-turbo-0613), G4 (gpt-4-0613). }
    \label{tab:automatic_metric}
\end{table}

\section{Supplementary Section 4}
\label{sec:supp_sec4}

In this section, we provide additional details of Section 4 in the main paper.

\subsection{Details for Proposer}

We ran each proposer for 3 rounds. For each round, we sample 20 examples per set and generate 10 hypotheses. 

\begin{figure}[htbp]
    \centering
    \includegraphics[width=\linewidth]{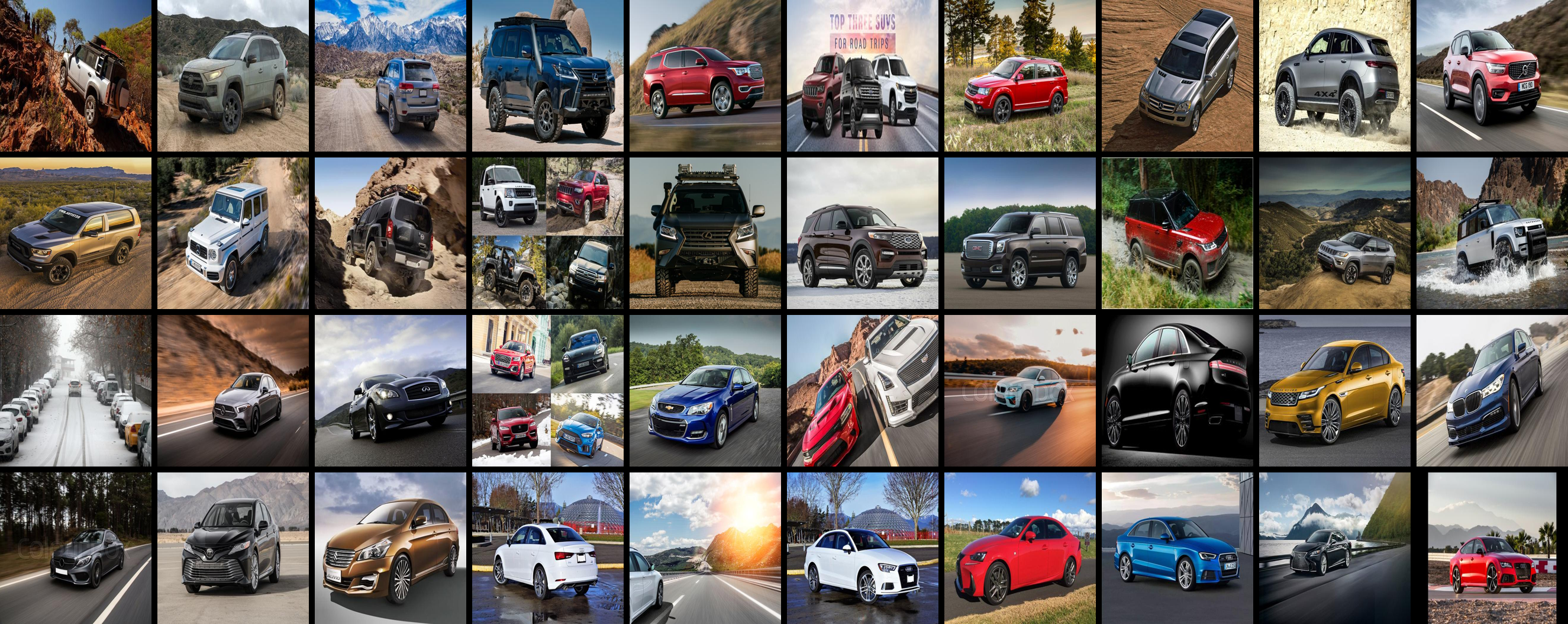}
    \caption{Example input to the image-based proposer. We arrange 20+20 input images into a single 4-row, 10-column gridded image.}
    \label{supp_fig:imgrid_example}
\end{figure}

\begin{figure*}[!tb]
    \centering
    \includegraphics[width=0.98\textwidth, trim={0cm 0cm 0cm 0cm},clip]{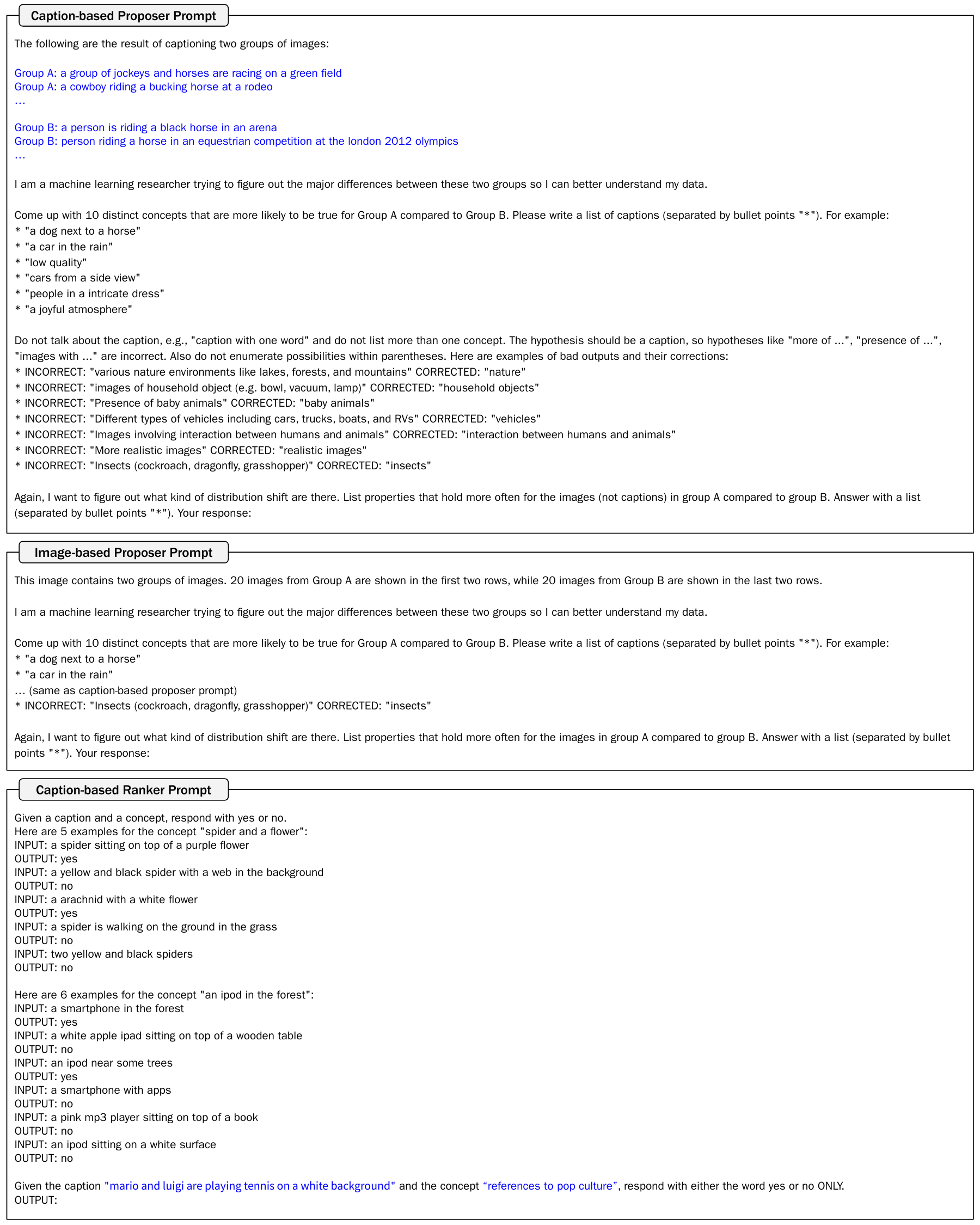}
    \caption{Prompt for caption-based proposer (top), image-based proposer (middle), and caption-based ranker (bottom). Input-dependent texts are colored in blue. }
    \label{supp_fig:supp_sec4_prompt}
\end{figure*}

\paragraph{Image-based Proposer.} We provide an example input of the gridded image in~\autoref{supp_fig:imgrid_example}. We feed the image and the prompt shown in~\autoref{supp_fig:supp_sec4_prompt} (middle) to LLaVA-1.5 to generate 10 hypotheses.

\paragraph{Feature-based Proposer.} To generate 10 hypotheses, we sample BLIP-2 10 times using top-p sampling given the subtracted embedding. 

\paragraph{Caption-based Proposer.} We generate captions for each image using BLIP-2 with the prompt ``Describe this image in detail.''. We apply a simple filtering of the captions, which removes any special tokens and captions simply repeating words. We feed the filtered captions and the prompt shown in~\autoref{supp_fig:supp_sec4_prompt} (top) to GPT-4 to generate 10 hypotheses.

\subsection{Details for Ranker}

\paragraph{Image-based Ranker.} Given each hypothesis, we prompt LLaVA-1.5 with the image and the prompt ``Does this image contain \{hypothesis\}?''.

\paragraph{Caption-based Ranker.} Given each hypothesis, we prompt Vicuna-1.5 with the image caption and hypothesis using the prompt shown in~\autoref{supp_fig:supp_sec4_prompt} (bottom).

\paragraph{Feature-based Ranker.} We use the OpenCLIP model ViT-bigG-14 trained on laion2b\_s39b\_b160k.

\subsection{Different Ranking Metrics} 

\autoref{tab:rebuttal_results} shows the results of several different ranking metrics using the CLIP similarity scores on VisDiffBench. We see that AUROC produces the most consistent highest performing results, but other metrics such as p-value and the difference in means also produce promising results.

\begin{table}[!tb]
    \centering
    \setlength{\tabcolsep}{4pt}
    \resizebox{\linewidth}{!}{  
    \begin{tabular}{l|cc|cc|cc|cc}
        \toprule
        \multirow{2}{*}{\textbf{Metric}} & \multicolumn{2}{c|}{\textbf{Easy}} & \multicolumn{2}{c|}{\textbf{Medium}} & \multicolumn{2}{c|}{\textbf{Hard}} & \multicolumn{2}{c}{\textbf{IN-R/*}}  \\ 
        & A1 & A5 & A1 & A5 & A1 & A5 & A1 & A5 \\ \midrule
        AUROC & \textbf{0.88} & \textbf{0.99} & \textbf{0.75} & \textbf{0.86} & \textbf{0.61} & \textbf{0.80} & 0.78 & \textbf{0.96} \\
        p-value  & 0.83 & \textbf{0.99} & 0.74 & \textbf{0.86} & 0.58 & 0.77 & \textbf{0.81} & 0.95 \\
        diff. in means & 0.83 & 0.98 & 0.69 & 0.84 & 0.60 & 0.76 & 0.76 & 0.92  \\
        \bottomrule
    \end{tabular}
    }
    \caption{\textbf{\benchmark{} results using different ranking metrics based on CLIP similarity scores.} We use the caption-based proposer. A1 \& A5 are Acc@1 \& Acc@5. }
    \label{tab:rebuttal_results}
\end{table}

\section{Supplementary Section 5}
\label{sec:supp_sec5}

In this section, we provide additional details of Section 5 in the main paper. We ablate various design choices of \method{}.

\begin{table*}[htbp]
    \small
    \centering
    \begin{tabular}{ll|cc|cc|cc|cc}
        \toprule
        \multirow{2}{*}{\textbf{Proposer}} & \multirow{2}{*}{\textbf{Ranker}} & \multicolumn{2}{c|}{\textbf{PIS-Easy}} & \multicolumn{2}{c|}{\textbf{PIS-Medium}} & \multicolumn{2}{c|}{\textbf{PIS-Hard}} & \multicolumn{2}{c}{\textbf{ImageNet-R/*}}  \\ 
        & & Acc@1 & Acc@5 & Acc@1 & Acc@5 & Acc@1 & Acc@5 & Acc@1 & Acc@5 \\ \midrule
        GPT-4 on BLIP-2 Captions & CLIP & \textbf{0.88} & \textbf{0.99} & \textbf{0.75} & \textbf{0.86} & \textbf{0.61} & \textbf{0.80} & 0.78 & \textbf{0.96} \\
        GPT-4 on LLaVA-1.5 Captions & CLIP & \textbf{0.89} & \textbf{0.98} & \textbf{0.73} & \textbf{0.85} & 0.51 & 0.70 & \textbf{0.84} & 0.93  \\
        GPT-3.5 on BLIP-2 Captions & CLIP & 0.81 & 0.95 & 0.67 & \textbf{0.87} & \textbf{0.60} & 0.76 & \textbf{0.85} & \textbf{0.96} \\
        \bottomrule
    \end{tabular}
    \caption{Results on VisDiffBench with different captions and language models. We bold any numbers within 0.02.}
    \label{supp_tab:results}
\end{table*}

\begin{figure*}[!tb]
    \centering
    \includegraphics[width=0.49\linewidth]{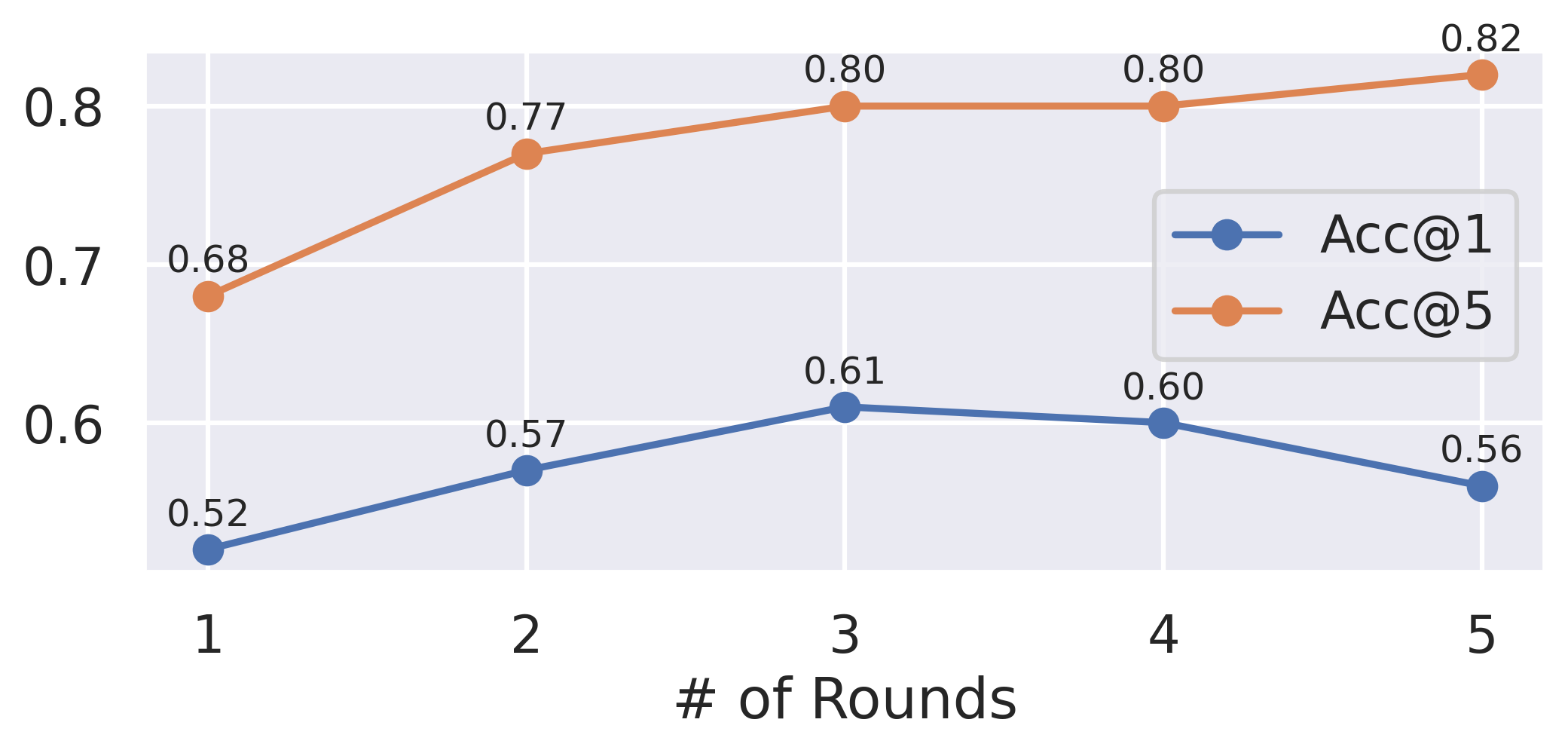}
    \includegraphics[width=0.49\linewidth]{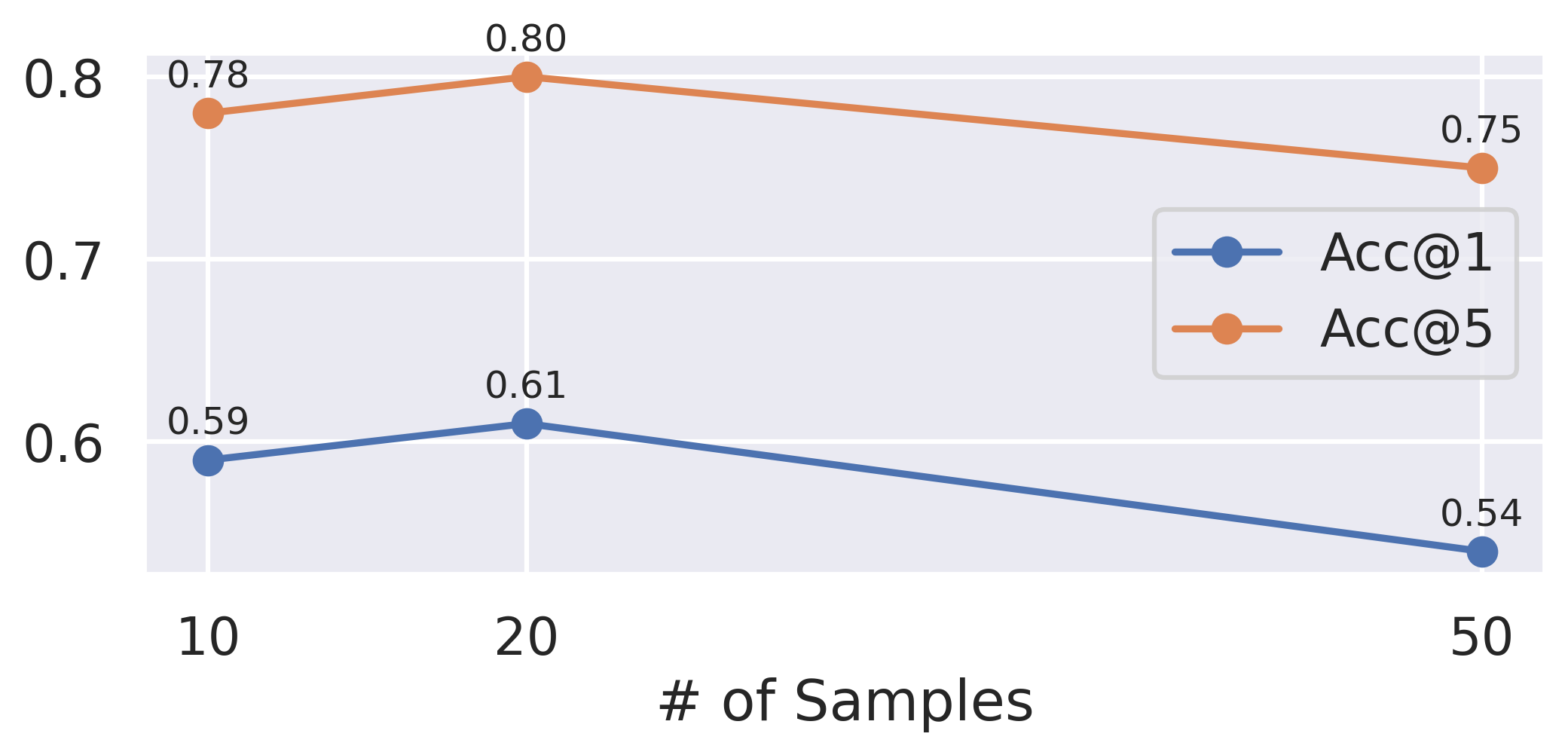}
    \caption{Analysis of the number of rounds (left) and number of samples (right) for the proposer on 50 \ourbenchmark{}-Hard sets. 3 rounds and 20 samples are the best in terms of performance and efficiency.}
    \label{supp_fig:round_sample_analysis}
\end{figure*}

\subsection{Caption Styles}

Given that our leading proposer is caption-based, it naturally raises the question of how captions derived from vision language models influence performance. We conducted a comparative analysis of captions generated by two state-of-the-art vision language models: BLIP-2 and LLaVA-1.5. Notably, compared to BLIP-2, LLaVA-1.5 has been instruction-tuned and can produce captions that are much longer with detailed information. The average caption length for LLaVA is around 391 characters compared to BLIP-2's 41 characters. As shown in~\autoref{supp_tab:results}, despite the clear disparity between these two captions, the algorithm achieves similar performances. This suggests that language models possess a robust inductive reasoning ability that allows them to discern the most notable differences in language. BLIP-2's captions, being marginally superior, could be attributed to their shortness and conciseness.

\subsection{Language Models} 

We compared GPT-4 with GPT-3.5 in~\autoref{supp_tab:results} to assess how different language models affect the caption-based proposer. While both models achieve strong performances on \benchmark{}, GPT-4 outperforms GPT-3.5 in most cases, demonstrating that the stronger reasoning capability of language models is important to accomplish the \task{} task.

\subsection{Sampling Rounds}

The proposer's generated differences rely on the random samples drawn from each image set; thus, extensive sampling is paramount to capture all the differences. Our ablation studies presented in \autoref{supp_fig:round_sample_analysis} (left), conducted on the 
\ourbenchmark{} hard subset, suggest that an increase in sampling iterations typically correlates with enhanced performance. However, a point of diminishing returns is observed beyond three rounds of proposals. In this paper, we standardize the experiments based on three rounds of proposal generation.

\subsection{Number of Sampled Examples}

Inputting more samples from \seta{} and \setb{} into the proposer may not be advantageous, as a long context could result in information getting lost in the middle~\cite{lost_in_the_middle,shi2023large}. Results shown in \autoref{supp_fig:round_sample_analysis} (right) reflect this, as inputting more captions to the large language models sees performance benefits up to 20 captions, at which point performance degrades.

\subsection{Necessity of Ranker}

Since the proposer may already generate and sort the most salient difference descriptions based on the sampled images, we conduct ablations to understand whether the ranker is necessary. We observe that, on \ourbenchmark{} hard subset, \method{} achieves 0.54 Acc@1 and 0.68 Acc@5 without ranker, which is much lower than the 0.61 Acc@1 and 0.80 Acc@5 with ranker, demonstrating the necessity of the ranker.

\section{Supplementary Section 6}
\label{sec:supp_sec6}

In this section, we provide additional details of Section 6 in the main paper.

\subsection{Comparing ImageNetV2 with ImageNet}
\label{sec:supp_imagenet_v2}

\begin{figure*}
     \centering
     \begin{subfigure}[htbp]{\textwidth}
         \centering
         \includegraphics[width=0.85\textwidth]{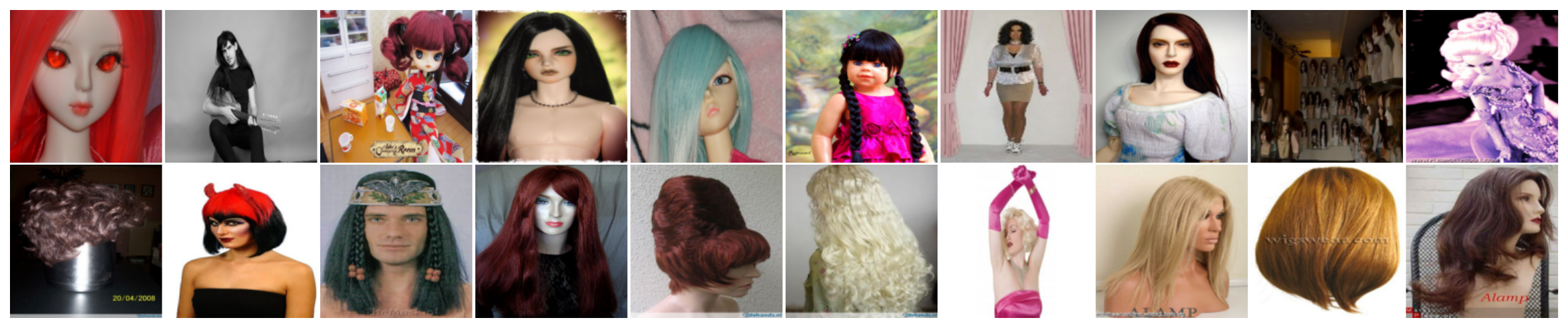}
         \caption{Wig  \hspace{0.5em} Diffs: ``Close up views of dolls'', ``Japanese style dolls'', ``Images including Barbie dolls''}
     \end{subfigure}

    \begin{subfigure}[htbp]{\textwidth}
         \centering
         \includegraphics[width=0.85\textwidth]{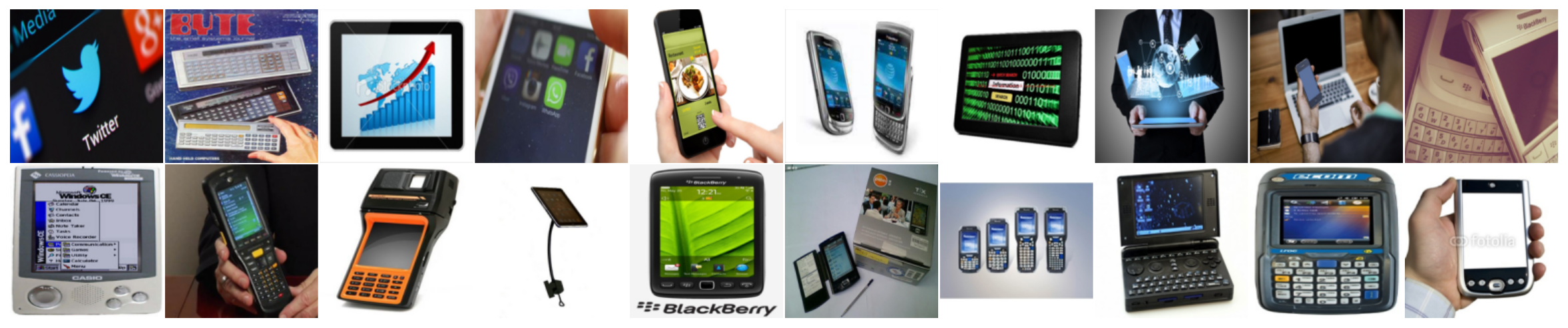}
         \caption{Hand-held Computer  \hspace{0.5em} Diffs: ``Apps like Twitter and Whatsapp'', ``Digital devices with green screen'', ``Interconnection between laptop and smart phone''}
     \end{subfigure}

     \begin{subfigure}[htbp]{\textwidth}
         \centering
         \includegraphics[width=0.85\textwidth]{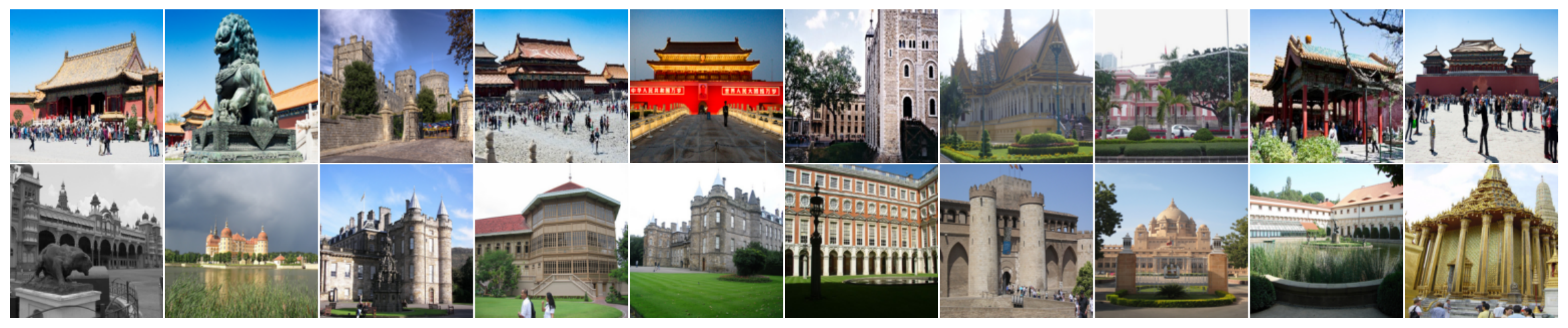}
         \caption{Palace \hspace{0.5em} Diffs: ``East Asian architecture'', ``Images featuring the Forbidden City in Beijing'', ``Images including red buildings with Chinese writing''}
     \end{subfigure}

     \begin{subfigure}[htbp]{\textwidth}
         \centering
         \includegraphics[width=0.85\textwidth]{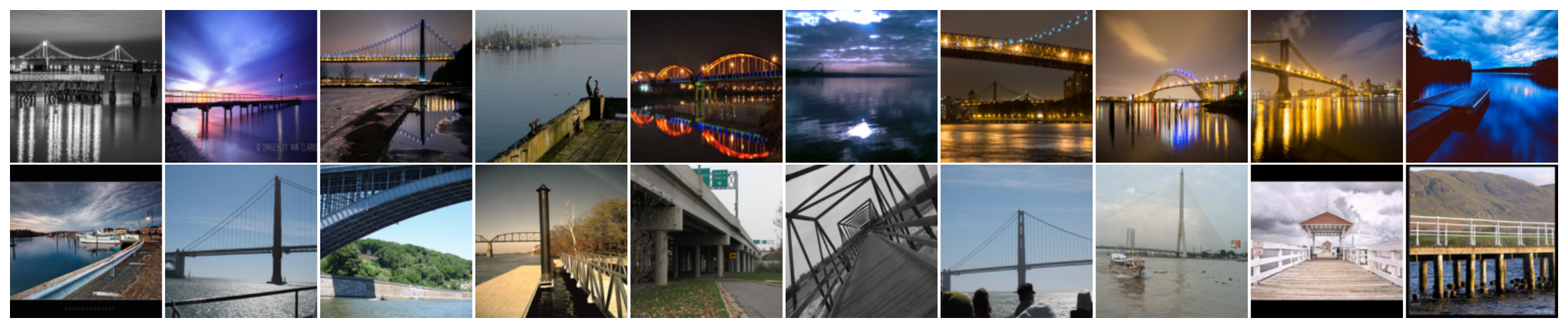}
         \caption{Pier \hspace{0.5em} Diffs: ``Body of water at night'',``Urban night skyline'', ``Long exposure shots''}
     \end{subfigure}

     \begin{subfigure}[htbp]{\textwidth}
         \centering
         \includegraphics[width=0.85\textwidth]{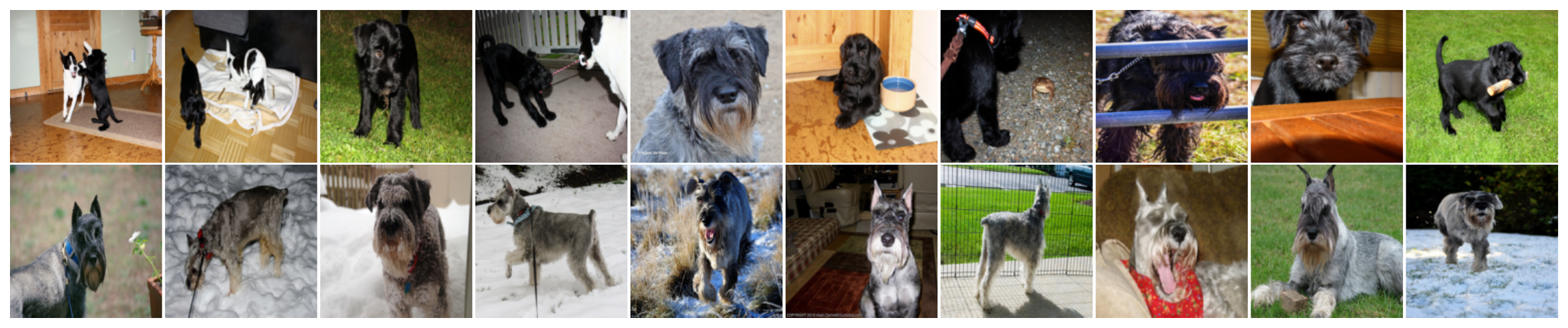}
         \caption{Schnauzer \hspace{0.5em} Diffs: ``Black dogs in different settings'', ``Terrier puppies with objects'', ``Interaction with different objects''}
     \end{subfigure}

     \begin{subfigure}[htbp]{\textwidth}
         \centering
         \includegraphics[width=0.85\textwidth]{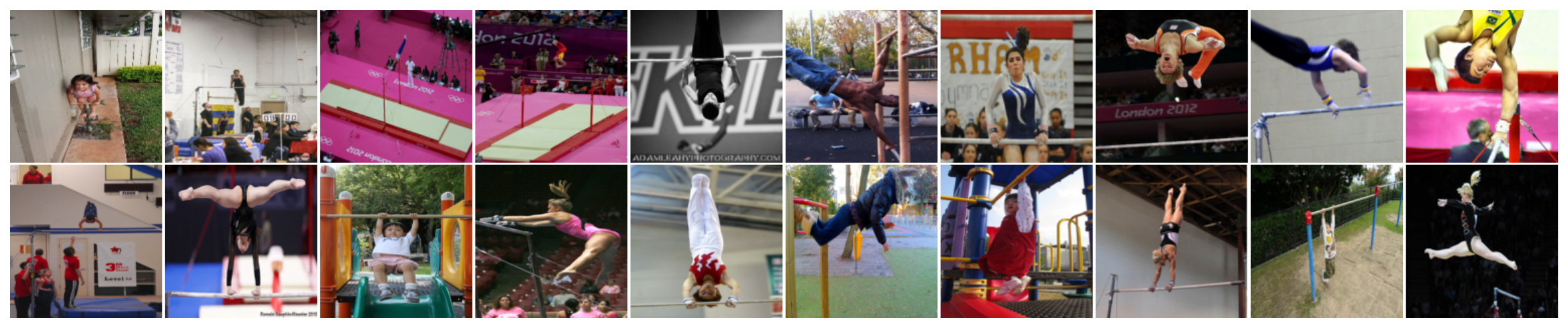}
         \caption{Horizontal Bar \hspace{0.5em} Diffs: ``Men's gymnastics events'', ``London 2012 Olympics'', ``Gymnastics event in 2013''}
     \end{subfigure}
     
    \caption{ImageNetV2 vs. ImageNet. All V2 images are shown in the first row while V1 images are shown in the second row. We show the class name and top 3 difference descriptions generated by \method{}. }
    \label{supp_fig:imagenet_v2_examples}
\end{figure*}

\paragraph{Per-class visualizations.} Along with the ``Dinner Table'' example shown in Figure 1, we provide other per-class differences with the highest difference scores in~\autoref{supp_fig:imagenet_v2_examples}. These examples clearly reveal salient differences between ImageNetV2 and ImageNet. Moreover, we observe time differences between these two datasets, as ImageNetV2 images contain Twitter and WhatsApp in the ``Hand-held Computer'' class and London 2012 Oylmpics in the ``Horizontal Bar'' class.

\paragraph{ImageNetV2 metadata analysis.} To get more precise statistics on when the ImageNetV2 images were collected, we analyzed the metadata of each image, which reports the minimum and maximum upload date of that image to Flickr. We find that 72\% images were uploaded between 2012 and 2013, and 28\% were uploaded between 2013 and 2014. This is different from ImageNet images that were all collected on or before 2010.

\subsection{Comparing Behaviors of CLIP and ResNet}
\label{sec:modeldiff_supp}

\begin{figure*}
     \centering
     \begin{subfigure}[htbp]{\textwidth}
         \centering
         \includegraphics[width=0.87\textwidth]{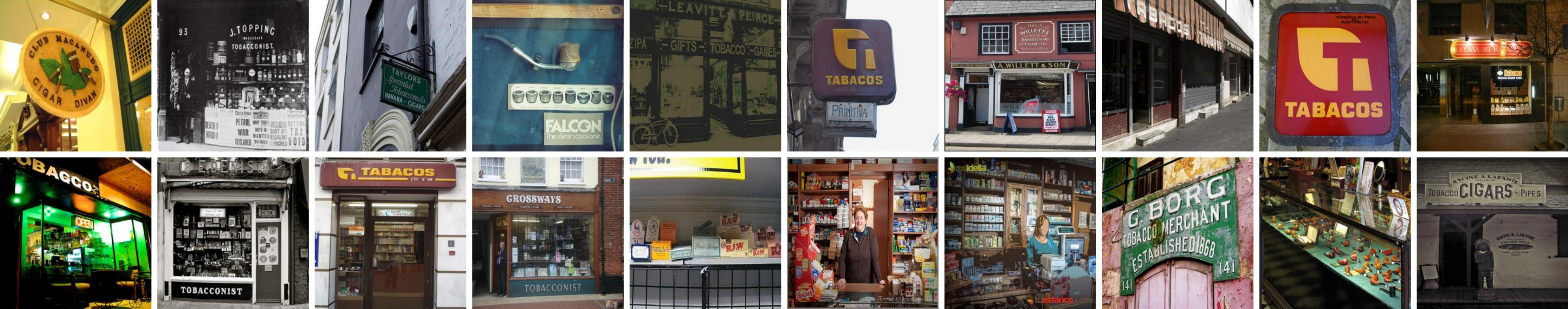}
         \caption{Tobacco Shop  \hspace{0.5em} Diffs: ``Sign hanging from the side of a building'',
         ``Falcon images'',``Presence of street signs''}
     \end{subfigure}

    \begin{subfigure}[htbp]{\textwidth}
         \centering
         \includegraphics[width=0.87\textwidth]{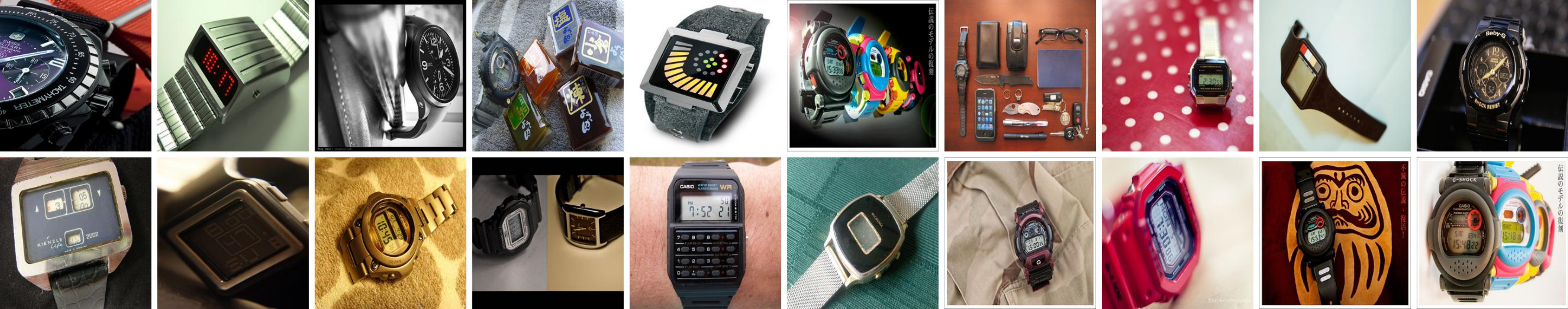}
         \caption{Digital Watch \hspace{0.5em} Diffs: ``Watches displayed in a group'',``Arrangement of multiple watches'',``Watches with colored straps''}
     \end{subfigure}

     \begin{subfigure}[htbp]{\textwidth}
         \centering
         \includegraphics[width=0.87\textwidth]{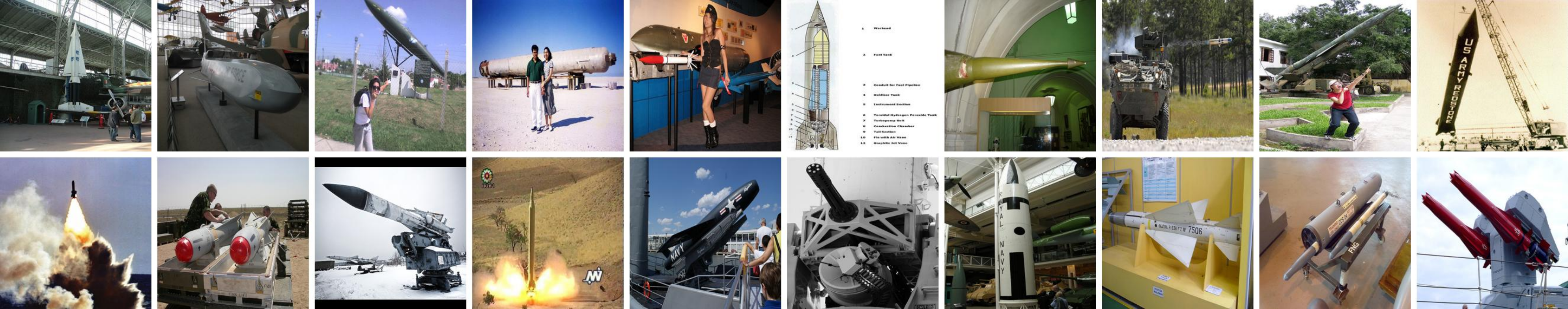}
         \caption{Missile \hspace{0.5em} Diffs: ``People posing with large missiles'',``people posing with rockets'',``missiles on display''}
     \end{subfigure}

     \begin{subfigure}[htbp]{\textwidth}
         \centering
         \includegraphics[width=0.87\textwidth]{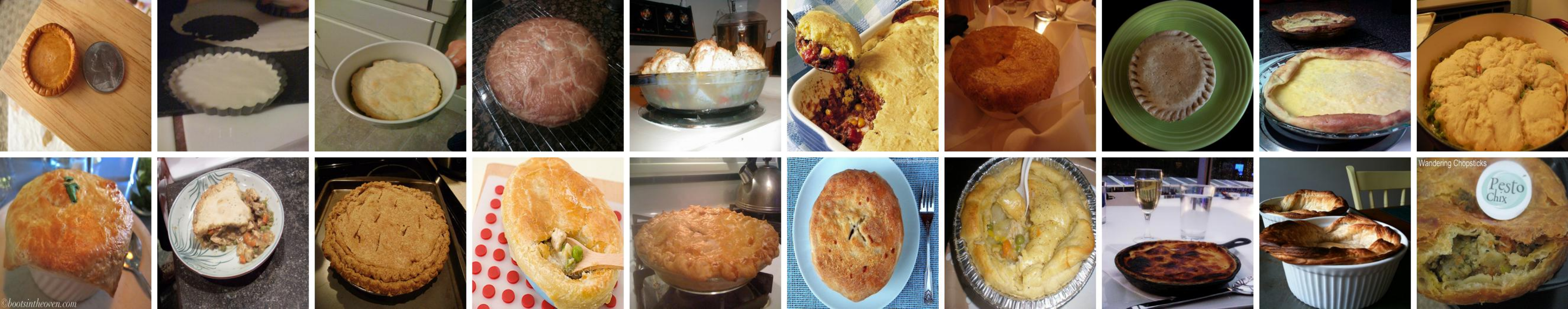}
         \caption{Pot Pie \hspace{0.5em} Diffs: ``Comparison of food size to coins'',``Utilization of cutting board'',``Inclusion of beverages''}
     \end{subfigure}

     \begin{subfigure}[htbp]{\textwidth}
         \centering
         \includegraphics[width=0.87\textwidth]{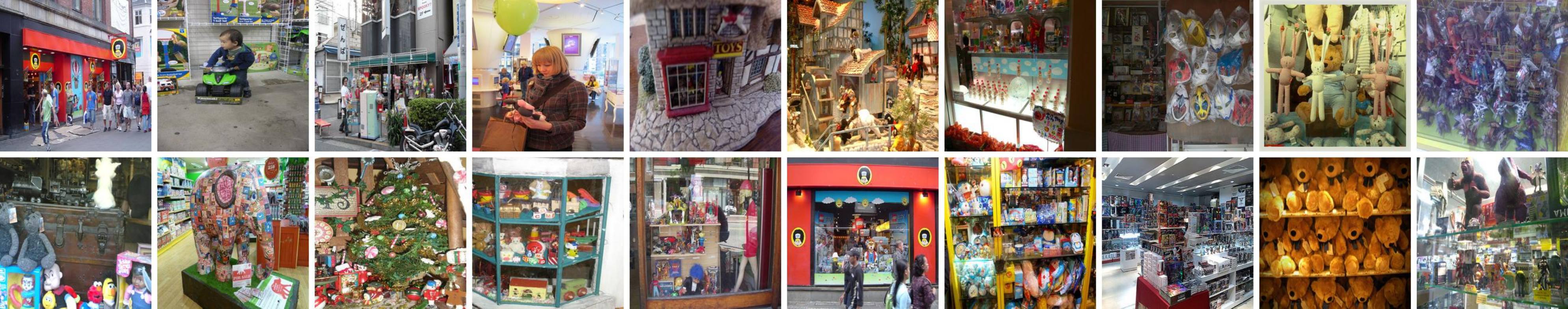}
         \caption{Toy Shop \hspace{0.5em} Diffs: ``People shopping in store'',``Female customer in store'',``Specific location based scene''}
     \end{subfigure}
     
    \caption{CLIP vs. ResNet. All CLIP correctly classified but ResNet incorrectly classified images are shown in the first row while other images are shown in the second row. We show the class name and top 3 difference descriptions generated by \method{}. }
    \label{supp_fig:modeldiff_examples}
\end{figure*}

\paragraph{Top Differences and Per-class visualizations.} We provide per-class differences where CLIP outperforms ResNet most in~\autoref{supp_fig:modeldiff_examples}. These examples clearly reveal salient differences between CLIP and ResNet, such as CLIP's robustness to label within images, object angle, and presence of people.

\subsection{Finding Failure Modes of ResNet}
\label{sec:resnet_failure_supp}

\paragraph{Model details.} We use the PyTorch pre-trained ResNet-50 and ResNet-101 models and the Huggingface ``facebook/detr-resnet-50'' object detector. 

\paragraph{Top differences.} The top 5 difference descriptions from \method{} were ``humanized object items'', ``people interacting with objects'', ``electronics and appliances'', ``objects or people in a marketplace setting'', and ``household objects in unusual placement''. 

\subsection{Comparing Versions of Stable Diffusion} 
\label{sec:diffusion_supp}

\begin{figure*}[htbp]
    \centering
    \includegraphics[width=\textwidth]{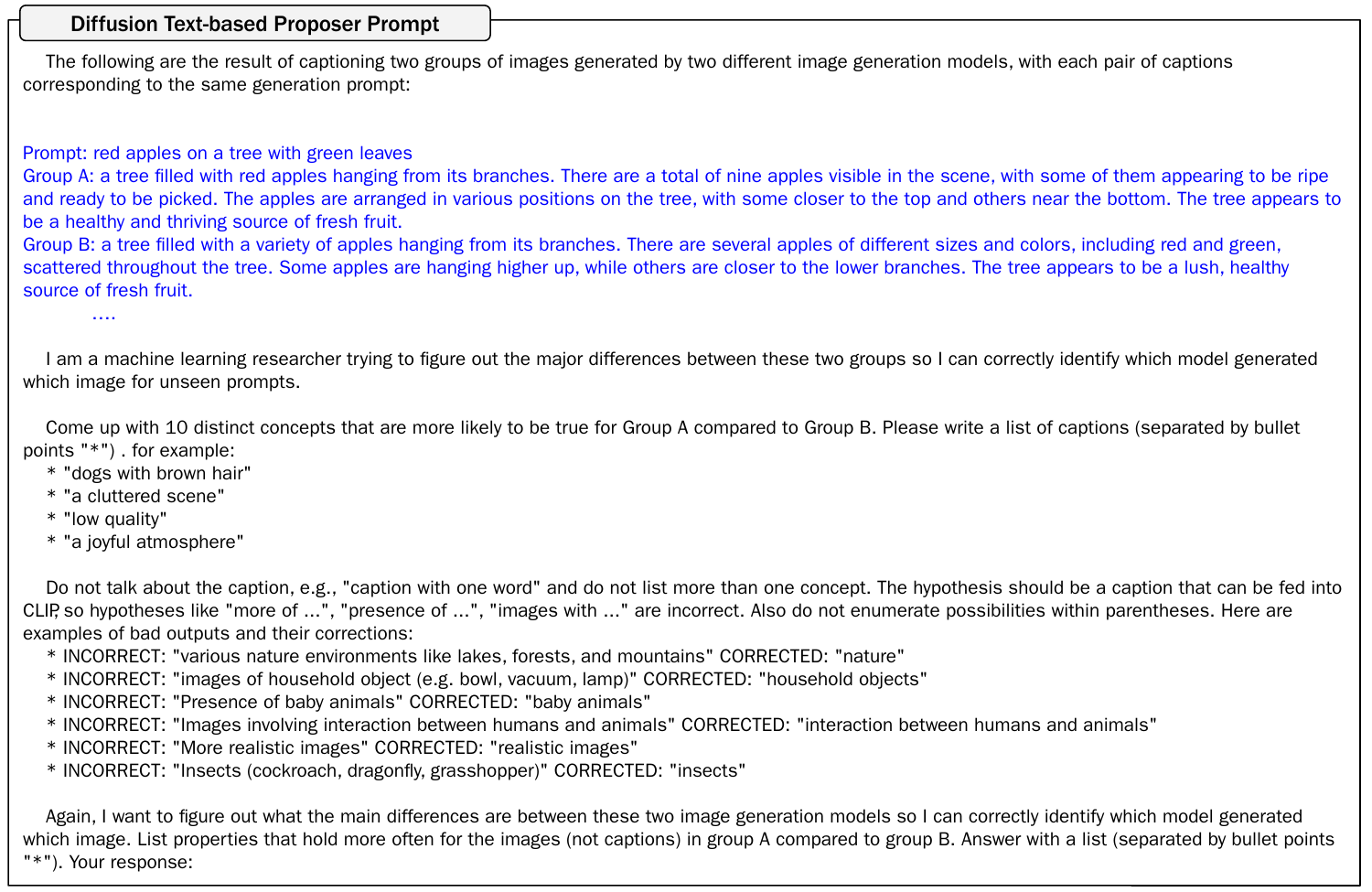}
    \caption{Modified proposer's prompt for StableDiffusion analysis.}
    \label{supp_fig:diffusion_prompt}
\end{figure*}

\paragraph{Text-to-image generation details.} We use the Huggingface models ``CompVis/stable-diffusion-v1-4'' and ``stabilityai/stable-diffusion-2-1'' with guidance of 7.5 and negative prompts ``bad anatomy, bad proportions, blurry, cloned face, cropped, deformed, dehydrated, disfigured, duplicate, error, extra arms, extra fingers, extra legs, extra limbs, fused fingers, gross proportions, jpeg artifacts, long neck, low quality, lowres, malformed limbs, missing arms, missing legs, morbid, mutated hands, mutation, mutilated, out of frame, poorly drawn face, poorly drawn hands, signature, text, too many fingers, ugly, username, watermark, worst quality''. 

\paragraph{\method{} details.} Unlike the previous applications, there exists a one-to-one mapping between \seta{} and \setb{} through the generation prompt. Therefore, we modify the subset sampling process to include the images generated from the same prompts and modify the proposer's prompt to include the generation prompts (\autoref{supp_fig:diffusion_prompt}). We used LLaVA-1.5 for captioning rather than BLIP-2 because we were particularly interested in the details of the images. 

\paragraph{Top differences.} Top 5 differences are shown in~\autoref{supp_tab:diffusion_hypotheses}.

\begin{table}[htbp]
    \scriptsize
    \centering
    \begin{tabular}{l|cc}
    \toprule
        & \multicolumn{2}{c}{\textbf{AUROC}} \\
        \textbf{More True for SDv2} &  \textbf{Parti}  & \textbf{DiffDB}\\
        \midrule
         colorful and dynamic collages of shapes or items & 0.70 & 0.71\\
         vibrant colors & 0.72 & 0.70 \\
         strong contrast in colors & 0.68 & 0.68\\
         reflective surfaces & 0.68 & 0.68\\
         artworks placed on stands or in frames & 0.64 & 0.66\\
         \bottomrule
    \end{tabular}
    \caption{Concepts more true for SDv2 than v1. Differences are proposed by running \method{} on PartiPrompts images. These differences obtain similar scores on the unseen DiffusionDB images, indicating that these differences generalize to various prompts. }
    \label{supp_tab:diffusion_hypotheses}
\end{table}

\begin{figure*}[htbp]
    \centering
    \includegraphics[width=\textwidth]{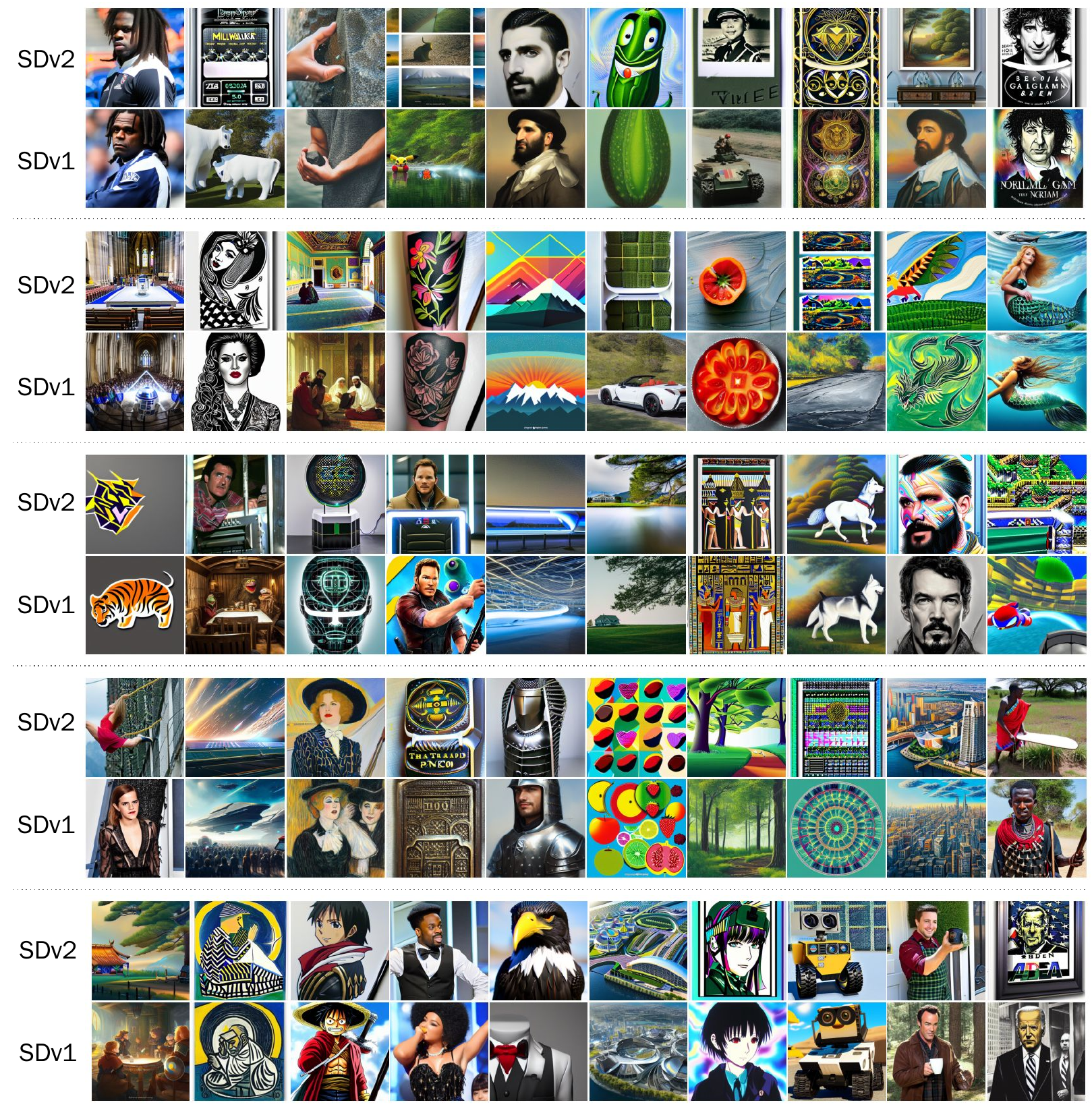}
    \caption{Randomly sampled images generated from SDv2 and v1 using DiffusionDB prompts. }
    \label{supp_fig:diffusion_random_samples}
\end{figure*}

\paragraph{Visualizations.} We provide 50 random samples of SDv2 and SDv1 images generated with DiffusionDB prompts in~\autoref{supp_fig:diffusion_random_samples}. These examples clearly verify that SDv2-generated images contain more vibrant contrasting colors and artwork or objects in frames or stands.

\paragraph{Edge analysis.} One interesting finding from \method{} is that SDv2 generated images contain more image frames than SDv1, such as a white border characterized by thick, straight lines spanning much of the image. To quantify this, we employed a Canny edge detector and searched for straight white lines in the images, with a thickness ranging from 5 to 20 pixels and a length exceeding 300 pixels (given the image size is 512x512). Applying this analysis to DiffusionDB images revealed that 13.6\% of SDv2 images exhibited such lines, as opposed to only 5.4\% from SDv1. This statistic provides additional evidence for such difference.

\subsection{Memorable Images} 
\label{sec:memorability_supp}

\paragraph{Top differences.} The top 25 difference descriptions generated by \method{} are presented in~\autoref{supp_tab:memorable_hypotheses}.

\begin{table}[htbp]
    \centering
    \scriptsize
    \begin{tabular}{l|p{0.75\linewidth}}
\toprule
Memorable & 
close-up of individual people,
use of accessories or personal items,
tattoos on human skin,
close-up on individuals,
humorous or funny elements,
artistic or unnaturally altered human features,
humorous elements,
detailed description of tattoos,
fashion and personal grooming activities,
pop culture references,
collectibles or hobbies,
light-hearted or humorous elements,
themed costumes or quirky outfits,
animated or cartoonish characters,
emphasis on fashion or personal style,
close-up of objects or body parts,
close-up facial expressions,
unconventional use of everyday items,
images with a playful or humorous element,
focus on specific body parts,
silly or humorous elements,
people in casual or humorous situations,
detailed description of attire,
quirky and amusing objects,
humorous or playful expressions \\
\midrule
Forgettable & Sunsets and sunrises,
  serene beach settings,
  sunset or nighttime scenes,
  agricultural fields,
  clear daytime outdoor settings,
  landscapes with water bodies,
  images captured during different times of day and night,
  Beautiful skies or sunsets,
  abandoned or isolated structures,
  natural elements like trees and water,
  urban cityscapes,
  urban cityscapes at night,
  various weather conditions,
  Afar shots of buildings or architectural structures,
  outdoor landscapes,
  cityscapes,
  Cityscapes and urban environments,
  Scenic outdoor landscapes,
  landscapes with mountains,
  Picturesque mountain views,
  expansive outdoor landscapes,
  Scenic landscapes or nature settings,
  Serene and tranquil environments,
  scenic landscapes,
  scenes with a serene and peaceful atmosphere \\
\bottomrule
    \end{tabular}
    \caption{Top 25 differences for memorable and forgettable images.}
    \label{supp_tab:memorable_hypotheses}
\end{table}

\paragraph{Classification analysis.} To validate whether the generated differences for memorable and forgettable images make sense, we use CLIP to classify each image in the LaMem dataset to these 25+25 differences and then assign the label ``forgettable'' or ``memorable'' based on where the difference is from. For example, if an image has the highest cosine similarity with ``close-up of individual people'', we assign its label as ``memorable''. We observe a 89.8\% accuracy on the LaMem test set, demonstrating that these differences provide strong evidence to classify whether images are memorable or forgettable. 

\section{Failure Cases and Limitations}
\label{sec:supp_limitations}

\begin{figure*}[!tb]
     \centering
         \includegraphics[width=0.92\textwidth]{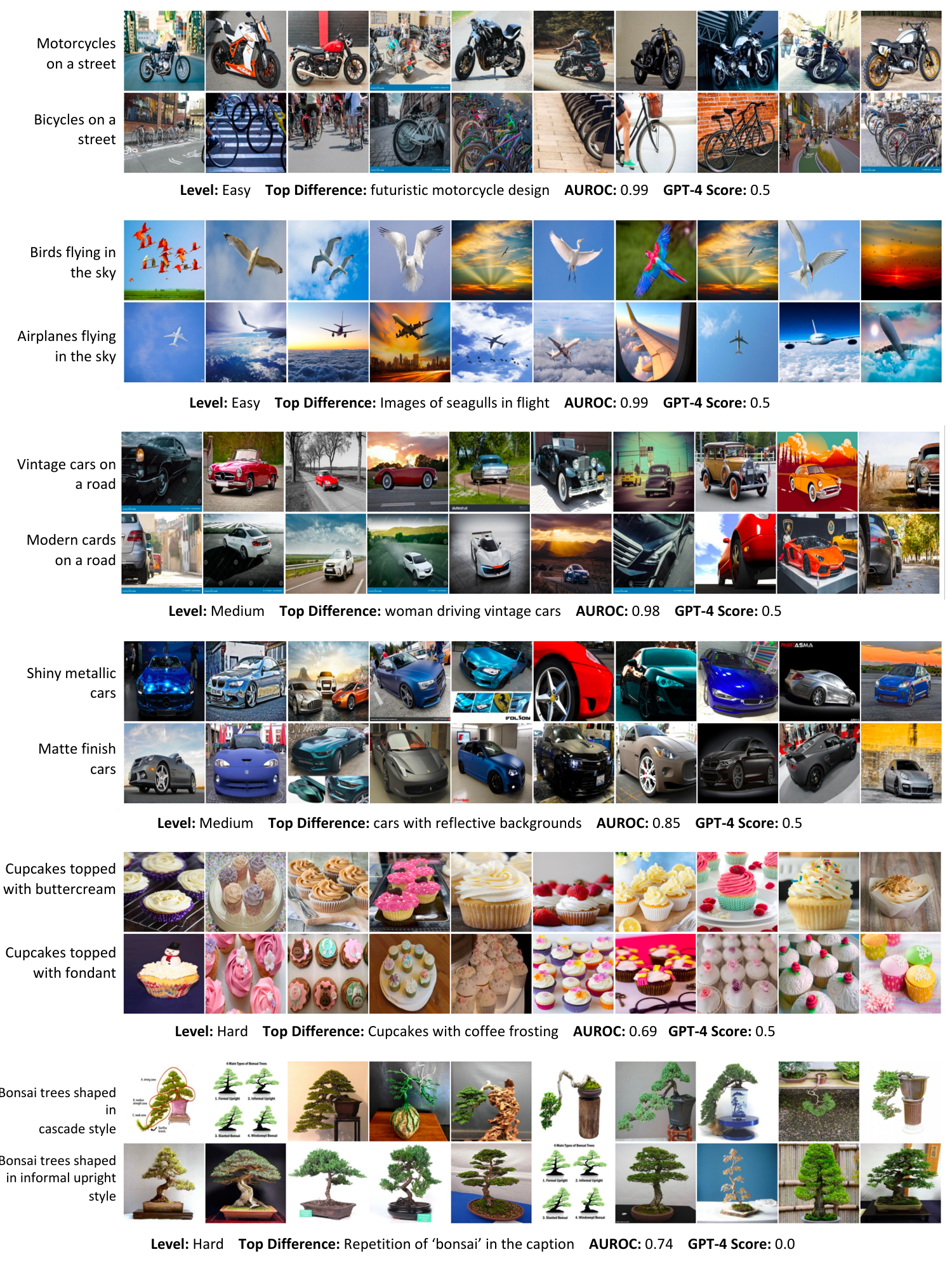}
        \caption{\ourbenchmark{} examples where \method{} fails. We show the ground-truth difference, top difference predicted by \method{}, AUROC score output by the ranker, and evaluation of the predicted difference by GPT-4.}
        \label{supp_fig:failed_visdiff_examples}
\end{figure*}

In this section, we summarize the failure cases and limitations of \method{} algorithm.

\subsection{Caption-based Proposer}

While our evaluation in the main paper shows that the caption-based proposer outperforms other counterparts by a large margin, translating images to captions may lead to information loss. For example, as shown in~\autoref{supp_fig:failed_visdiff_examples}, fine-grained differences between groups ``Cupcakes topped with buttercream'' and ``Cupcakes topped with fondant'' is overlooked due to generic captions. We expect using captioning prompts tailored to the application domain can mitigate this issue.

Furthermore, despite providing task context and several in-context examples, we noted instances where GPT-4 predominantly focused on the captions rather than the underlying high-level visual concepts. A frequent error involves generating concepts related more to the caption than the image, such as ``repetition of 'bonsai' in the caption,'' as illustrated in~\autoref{supp_fig:failed_visdiff_examples}. We anticipate that this issue will diminish as LLMs' instruction-following ability improves.

\subsection{Feature-based Ranker} 

Several of \method{}'s ranker failure cases stem from biases and limitations in CLIP. For example, nuanced differences such as ``a plant to the left of the couch'' are often assigned lower rankings because CLIP struggles with precise location details, and minor variations in phrasing can lead to significant differences in similarity scores.

Additionally, using AUROC on cosine similarities as a ranking metric is sensitive to outliers in cosine similarity scores. In practice, we have noticed that outliers can cause very specific difference descriptions to be scored higher than more general differences. For instance, as shown in~\autoref{supp_fig:failed_visdiff_examples}, with \seta{} being ``Birds flying in the sky'' and \setb{} ``Airplanes flying in the sky,'' the hypothesis ``Images of seagulls in flight'' received a higher AUROC score than the more broadly applicable ``birds in flight''.

\subsection{LLM-based Evaluation} 

As demonstrated in the main paper, large language models generally align well with human evaluations. However, there are instances where they fail to accurately score differences against the ground truth descriptions. An example from \benchmark{} involves the description ``Green apples in a basket'' for \seta{} and ``Red apples in a basket'' for \setb{}. Here, the top hypothesis by \method{}, ``Green apples'' received a score of only 0.5 instead of the expected 1.0. These errors are expected to diminish as LLM improves.

\subsection{\benchmark{}} 

Most differences in \benchmark{} focus on objects, styles, and actions. Differences such as object position, size, or image quality are missing. Additionally, since \ourbenchmark{} is compiled by scraping images from the web, the datasets inevitably include noise. For instance, searching for ``a cat to the left of a dog'' often yields images with a cat on the right instead.

\subsection{Reliance on Large Pre-trained Models} 

Our approach is fundamentally based on large, pre-trained vision-language foundation models. These models' extensive capabilities make them adaptable for a variety of tasks. However, inherent biases and limitations in these models may be transferred to our method.  Additionally, these models might be confined to domains observed during pre-training, potentially limiting their applicability to novel domains, such as biomedical imaging. Nevertheless, we anticipate that rapid advancements in foundation model development will mitigate these issues, thereby enhancing our method's effectiveness.

\end{document}